\documentclass{article}

\usepackage{microtype}
\usepackage{graphicx}
\usepackage{subfig}
\usepackage{booktabs} 
\usepackage{arydshln}

\usepackage{hyperref}

\newcommand\numberthis{\addtocounter{equation}{1}\tag{\theequation}}

\usepackage[accepted]{icml2020}

\usepackage{url}
\usepackage{amsmath}
\usepackage{amssymb}
\usepackage{graphicx}
\usepackage{multirow}
\usepackage{cleveref}
\DeclareMathOperator*{\D}{\mathcal{D}}

\DeclareMathOperator*{\R}{\mathbb{R}}
\newcommand{\E}{\mathbb{E}}
\DeclareMathOperator*{\argmin}{argmin}

\DeclareMathAlphabet\mathbfcal{OMS}{cmsy}{b}{n}

\crefformat{footnote}{#2\footnotemark[#1]#3}

\icmltitlerunning{An end-to-end approach for the verification problem: learning the right distance}

\begin{document}

\twocolumn[
\icmltitle{An end-to-end approach for the verification problem: learning the right distance}

\icmlsetsymbol{equal}{*}

\begin{icmlauthorlist}
\icmlauthor{Jo\~ao Monteiro}{inrs,crim}
\icmlauthor{Isabela Albuquerque}{inrs}
\icmlauthor{Jahangir Alam}{inrs,crim}
\icmlauthor{R Devon Hjelm}{msr,mila}
\icmlauthor{Tiago Falk}{inrs}
\end{icmlauthorlist}

\icmlaffiliation{inrs}{INRS-EMT, Universit\'e du Qu\'ebec, Montreal, Canada.}
\icmlaffiliation{crim}{Centre de Recherche Informatique de Montr\'eal, Montreal, Canada.}
\icmlaffiliation{mila}{Quebec Artificial Intelligence Institute, Universit\'e de Montr\'eal, Montreal, Canada.}
\icmlaffiliation{msr}{Microsoft Research.}

\icmlcorrespondingauthor{Jo\~ao Monteiro}{joao.monteiro@emt.inrs.ca}

\icmlkeywords{Verification, Metric learning, Contrastive Estimation}

\vskip 0.3in
]



\printAffiliationsAndNotice{}  

\begin{abstract}

In this contribution, we augment the metric learning setting by introducing a parametric pseudo-distance, trained jointly with the encoder. Several interpretations are thus drawn for the learned distance-like model's output. We first show it approximates a likelihood ratio which can be used for hypothesis tests, and that it further induces a large divergence across the joint distributions of pairs of examples from the same and from different classes. Evaluation is performed under the verification setting consisting of determining whether sets of examples belong to the same class, even if such classes are novel and were never presented to the model during training. Empirical evaluation shows such method defines an end-to-end approach for the verification problem, able to attain better performance than simple scorers such as those based on cosine similarity and further outperforming widely used downstream classifiers. We further observe training is much simplified under the proposed approach compared to metric learning with actual distances, requiring no complex scheme to harvest pairs of examples.
\end{abstract}

\section{Introduction}

Learning useful representations from high-dimensional data is one of the main goals of modern machine learning. However, doing so is generally a side effect of the solution of a pre-defined task, e.g., while learning the decision surface in a classification problem, inner layers of artificial neural networks are shown to make salient cues of input data which are discriminable. Moreover, in unsupervised settings, bottleneck layers of autoencoders as well as approximate posteriors from variational autoencoders have all been shown to embed relevant properties of input data which can be leveraged in downstream tasks. Rather than employing a neural network to solve some task and hope learned features are useful, approaches such as \textit{siamese networks} \cite{bromley1994signature}, which can be included in a set of approaches commonly referred to as \textit{Metric Learning}, have been introduced with the goal of explicitly inducing features holding desirable properties such as class separability. In this setting, an encoder is trained so as to minimize or maximize a \textit{distance} measured across pairs of encoded examples, depending on whether the examples within each pair belong to the same class or not, provided that class labels are available. Follow-up work leveraged this idea for several applications \cite{hadsell2006dimensionality,hoffer2015deep}, which include, for instance, the verification problem in biometrics, as is the case of FaceNet \cite{schroff2015facenet} and Deep-Speaker \cite{li2017deep}, which are used for face and speaker recognition, respectively. However, as pointed out in recent work \cite{schroff2015facenet,shi2016embedding,wu2017sampling,li2017deep,zhang2018text}, careful selection of training pairs is crucial to ensure a reasonable sample complexity during training given that most triplets of examples quickly reach the condition such that distances measured between pairs from the same class are smaller than those of the pairs from different classes. As such, developing efficient strategies for harvesting negative pairs with small distances throughout training becomes primordial.

In this contribution, we are concerned with the metric learning setting briefly described above, and more specifically, we turn our attention to its application to the verification problem, i.e., that of comparing data pairs and determining whether they belong to the same class. The verification problem arises in applications where comparisons of two small samples is required such as face/finger-print/voice verification \cite{reynolds2002overview}, image retrieval \cite{zhu2016deep, wu2017sampling}, and so on. At test time, inference is often performed to answer two types of questions: (i) Do two given examples belong to the same class? and (ii) Does a test example belong to a specific claimed class? And in both cases test examples might belong to classes never presented to the model during training. Current verification approaches are usually comprised of several components trained in a greedy manner \cite{kenny2013plda,snyder2018x}, and an end-to-end approach is still lacking.

Euclidean spaces will not, in general, be suitable for representing any desired type of structure expressed in the data (e.g. asymmetry \cite{pitis2020an} or hierarchy \cite{nickel2017poincare}). To avoid the need to select an adequate distance given every new problem we are faced with, as well as to deal with the training difficulties mentioned previously, we propose to augment the metric learning framework and jointly train an encoder (which embeds raw data into a lower dimensional space) and a (pseudo) distance model tailored to the problem of interest. An end-to-end approach for verification is then defined by employing such pseudo-distance to compute similarity scores. Both models together, parametrized by neural networks, define a (pseudo) metric space in which inference can be performed efficiently since now semantic properties of the data (e.g., discrepancies across classes) are encoded by scores. While doing so, we found several interpretations appear from such learned pseudo-distance, and it can be further interpreted as a likelihood ratio in a Neyman-Pearson hypothesis test, as well as an approximate divergence measure between the joint distributions of positive (same classes) and negative (different classes) pairs of examples. Moreover, even though we do not enforce models to satisfy properties of an actual metric\footnote{Symmetry, identity of indiscernibles, and triangle inequality.}, we empirically observe such properties to appear.

Our contributions can be summarized as follows:
\vspace{-0.4cm}
\begin{enumerate}
\item We propose an augmented metric learning framework where an encoder and a (pseudo) distance are trained jointly and define a (pseudo) metric space where inference can be done efficiently for verification.\vspace{-0.3cm}
\item We show that the optimal distance model for any fixed encoder yields the likelihood-ratio for a Neyman-Pearson hypothesis test, and it further induces a high Jensen-Shannon divergence between the joint distributions of positive and negative pairs.\vspace{-0.3cm}
\item The introduced setting is trained in an end-to-end fashion, and inference can be performed with a single forward pass, greatly simplifying current verification pipelines which involve several sub-components.\vspace{-0.3cm}
\item Evaluation on large scale verification tasks provides empirical evidence of the effectiveness in directly using outputs of the learned pseudo-distance for inference, outperforming commonly used downstream classifiers.\vspace{-0.3cm}
\end{enumerate}

The remainder of this paper is organized as follows: metric learning and the verification problem are discussed in Section \ref{sec:background}. The proposed method is presented in Section \ref{sec:method} along with our main guarantees, while empirical evaluation is presented in Section \ref{sec:evaluation}. Discussion and final remarks as well as future directions are presented in Section \ref{sec:discussion}.

\section{Background and related work}
\label{sec:background}

\subsection{Distance Metric Learning}

Being able to efficiently assess similarity across samples from data under analysis is a long standing problem within machine learning. Algorithms such as K-means, nearest-neighbors classifiers, and kernel methods generally rely on the selection of some similarity or distance measure able to encode semantic relationships present in high-dimensional data into real scores. Under this view, approaches commonly referred to as $\textit{Distance Metric Learning}$, introduced originally by \citet{xing2003distance}, try to learn a so-called Mahalanobis distance, which, given $x,y \in \R^n$, will have the form: $\sqrt{(x-y)^\intercal A(x-y)}$, where $A \in \R^{n \times n}$ is positive semidefinite. Several extensions of that setting were then introduced \cite{globerson2006metric,weinberger2009distance,ying2012distance}. 

\citet{shalev2004online}, for instance, proposed an online version of the algorithm in \cite{xing2003distance}, while an approach based on support vector machines was introduced in \cite{schultz2004learning} for learning $A$. \citet{davis2007information} provided an information-theoretic approach to solve for $A$ by minimizing the divergence between Gaussian distributions associated to the learned and the Euclidean distances, further showing such an approach to be equivalent to low-rank kernel learning \cite{kulis2006learning}. Similar distances have also been used in other settings, such as similarity scoring for contrastive learning \cite{oord2018representation,tian2019contrastive}. Besides the Mahalanobis distance, other forms of distance/similarity have been considered in recent work. In \cite{lanckriet2004learning}, for example, a kernel matrix is directly learned, implicitly defining a similarity function. In \cite{pitis2020an}, classes of neural networks are proposed to define pseudo-distances which satisfy the triangle inequality while not being necessarily symmetric.

For the particular case of Mahalanobis distance metric learning, one can show that $\exists \text{ } W : \sqrt{(x-y)^\intercal A(x-y)} = ||Wx-Wy||_2$ \cite{shalev2004online}, which means that there exists a linear projection of the data after which the Euclidean distance will correspond to the Mahalanobis distance on the original space. \citet{chopra2005learning} substituted the linear projection by a learned non-linear encoder $\mathcal{E}: R^D \rightarrow R^d$ so that $||\mathcal{E}(x)-\mathcal{E}(y)||_2$ yields a (non-Mahalanobis) distance measure between raw data points yielding useful properties. Follow-up work has extended such idea to several applications \cite{schroff2015facenet,shi2016embedding,li2017deep,zhang2018text}. One extra variation of $||Wx-Wy||_2$, besides the introduction of $\mathcal{E}$, is to switch the Euclidean distance $||\cdot||_2$ with an alternative better suited for the task of interest. That is the case in \cite{norouzi2012hamming}, where the Hamming distance is used over data encoded to a binary space. In \cite{courty2018learning}, in turn, the encoder is trained so that Euclidean distances in the encoded space approximate Wasserstein divergences, while \citet{nickel2018learning} employ a hyperbolic distance which is argued to be suitable for their particular use case.

Based on the covered literature, one can conclude that there are two different directions aimed at achieving a similar goal: \textit{learn to represent the data in a metric space where distances yield efficient inference mechanisms for various tasks}. While one corresponds to learning a meaningful distance or similarity from raw data, the other corresponds to, given a fixed distance metric, finding an encoding process yielding the desired metric space. Here, we propose an alternative to perform both these tasks {\it{simultaneously}}, i.e., jointly learn both the encoder and distance. Close to such an approach is the method discussed by \citet{garcia2019learning} where, similarly to our setting, both encoder and distance are trained, with the main differences lying in the facts that our method is fully end-to-end\footnote{What authors refer to as end-to-end requires pretraining an encoder in the metric learning setting with a standard distance.} while in their case training happens separately. Moreover, training of the distance model in that case is done by imitation learning of cosine similarities.

\subsection{The Verification Problem}

Given data instances $x \in \mathcal{X}$ such that each $x$ can be associated to a class label $y \in \mathcal{Y}$ through a labeling function $f:\mathcal{X} \rightarrow \mathcal{Y}$, we define a \textit{trial} as a pair of sets of examples $\{X_i,X_j\}$, provided that $f(x^k_i)=f(x^l_i) \text{ } \forall \text{ } k,l \in \{1,2,...,|X_i|\}^2$ and $f(x^k_j)=f(x^l_j) \text{ } \forall \text{ } k,l \in \{1,2,...,|X_j|\}^2$, so that we can assign class labels to such sets $X_m$ defining $f(X_m)=f(x_m) \text{ } \forall \text{ } x_m \in X_m$. The verification problem can be thus viewed as, given a trial $T_{i,j}=\{X_i,X_j\}$, deciding whether $f(X_i)=f(X_j)$, in which case we refer to $T$ as \textit{target trial}, or $f(X_i) \neq f(X_j)$ and the trial will be called \textit{non-target}.

\begin{figure}[h]
\centering
\includegraphics[width=0.49\textwidth, clip]{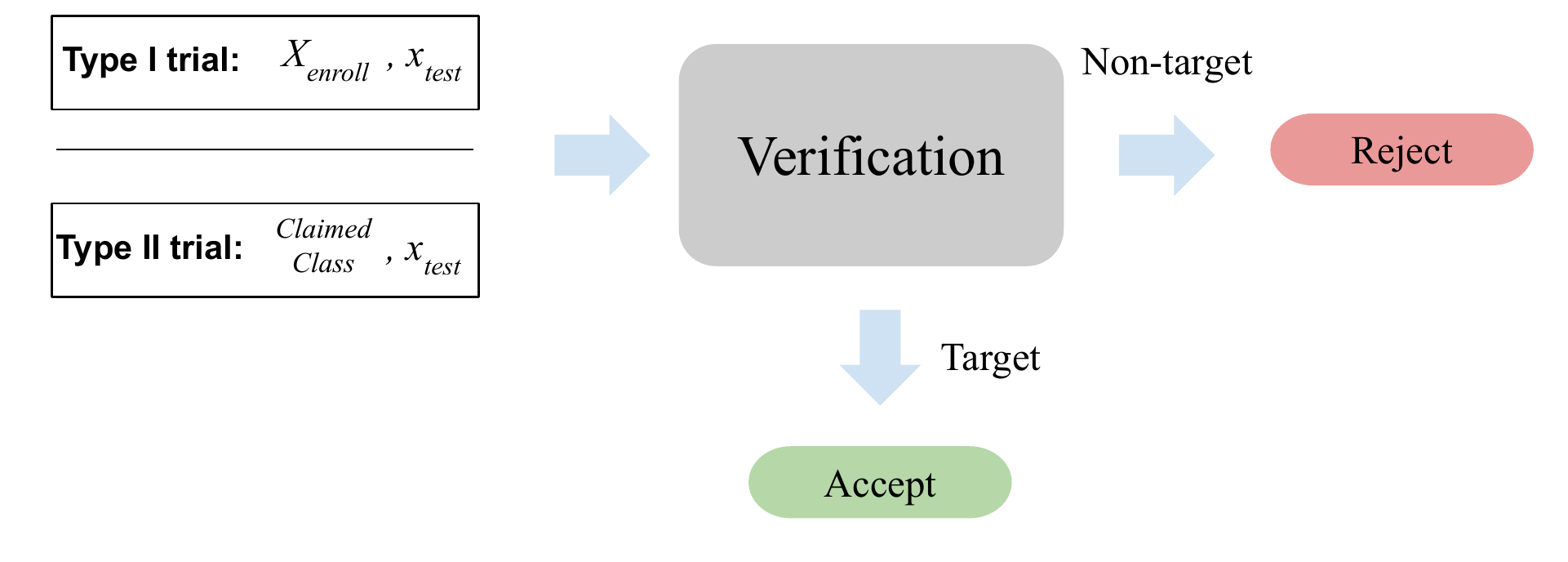}
\caption{The verification problem.}
\label{fig:verification_def}
\end{figure}

The verification problem is illustrated in Figure \ref{fig:verification_def}. We categorize trials into two types in accordance to practical instances of the verification problem: type I trials are those such that $X_i$ is referred to as enrollment sample, i.e., a set of data points representing a given class such as a gallery of face pictures from a given user in an access control application, while $X_j$ will correspond to a single example $x_{test}$ to be verified against the enrollment gallery. For the type II case, $X_i$ is simply a \textit{claim} corresponding to the class against which $x_{test}$ will be verified. Classes corresponding to examples within test trials might have never been presented to the model, and sets $X_i$ and $X_j$ are typically small ($<10$).

Under the Neyman-Pearson approach \cite{neyman1933ix}, verification is seen as a hypothesis test, where $H_0$ and $H_1$ correspond to the hypothesis such that $T$ is target or otherwise, respectively \cite{jiang2001bayesian}. The test is thus performed through the following likelihood ratio (LR):

\begin{equation}
LR = \frac{p(T|H_0)}{p(T|H_1)},
\label{eq:LR}
\end{equation}
where $p(T|H_0)$ and $p(T|H_1)$ correspond to models of target, and non-target (or impostor) trials. The decision is made by comparing $LR$ with a threshold $\tau$.

One can then explicitly approximate $LR$ through generative approaches \cite{deng2018speech}, which is commonly done using Gaussian mixture models. In that case, the denominator is usually defined as a universal background model (GMM-UBM, \citet{reynolds2000speaker}), meaning that it is trained on data from all available classes, while the numerator is a fine-tuned model on enrollment data so that, for trial $\{X_1,X_2\}$, $LR$ will be:

\begin{equation}
LR = \frac{p_{X_1}(X_2)}{p_{UBM}(X_2)} = \frac{p_{X_{Enroll}}(x_{test})}{p_{UBM}(x_{test})}.
\end{equation}

Alternatively, \citet{cumani2013pairwise} showed that discriminative settings, i.e., binary classifiers trained on top of data pairs to determine whether they belong to the same class, yielded likelihood ratios useful for verification. In their case, a binary SVM was trained on pairs of i-vectors \cite{dehak2010front} for automatic speaker verification. We build upon such discriminative setting, but with the difference that we learn an encoding process along with the discriminator (here represented as a distance model), and show it to yield likelihood ratios required for verification through contrastive estimation results. This is more general than the result in \cite{cumani2013pairwise}, which shows that there exists a generative classifier associated to each discriminator whose likelihood ratio matches the discriminator's output, requiring such classifier's assumptions to hold.


We remark that current verification approaches are composed of complex pipelines containing several components \cite{dehak2010front,kenny2013plda,snyder2018x}, including a pretrained data encoder, followed by a downstream classifier, such as probabilistic linear discriminant analysis (PLDA) \cite{ioffe2006probabilistic,prince2007probabilistic}, and score normalization \cite{auckenthaler2000score}, each contributing practical issues (e.g., cohort selection) to the overall system. This renders both training and testing of such systems difficult. The approach proposed herein is a step towards end-to-end verification, i.e., from data to scores via a single forward pass, thus simplifying inference.

\section{Learning pseudo metric spaces}
\label{sec:method}

We consider the setting where both an encoding mechanism, as well as some type of similarity or distance across data points are to be learned. Assume $\mathcal{E}: \R^D \rightarrow \R^d$ and $\D: \R^d \times \R^d \rightarrow (0,1)$ are deterministic mappings which will be referred to as encoder and distance model, respectively, and will be both parametrized by neural networks. Such entities resemble a metric-space, thus we will refer to it as \textit{pseudo metric space}. We empirically observed that introducing distance properties in $\D$, i.e., by constraining it to be symmetric and enforcing it to satisfy the triangle inequality, did not result in improved performance, yet rendered training unstable. However, since trained models are found to approximately behave as an actual distance, we make use of the analogy, but further provide alternative interpretations of $\D$'s outputs.

Data samples are such that $x \in \mathcal{X} \subset \R^D$, and $z = \mathcal{E}(x)$ represents embedded data in $\R^d$. It will be usually the case that $D \gg d$. Once more, each data example can be further assigned to one of $L$ class labels through a labeling function $f : \mathcal{X} \rightarrow \{1,...,L \}$. Moreover, we define positive and negative pairs of examples denoted by $+$ or $-$ superscripts such that $x^{+} = \{x_i, x_j\} \implies f(x_i)=f(x_j)$, as well as $x^{-} = \{x_i, x_j\} \implies f(x_i) \neq f(x_j)$. The same notation is employed in the embedding space so that $z^{+}=\mathcal{E}(x^{+})=\{\mathcal{E}(x_i),\mathcal{E}(x_j)\} \implies f(x_i)=f(x_j)$, and $z^{-}=\mathcal{E}(x^{-})=\{\mathcal{E}(x_i),\mathcal{E}(x_j)\} \implies f(x_i) \neq f(x_j)$. We will denote the sets of all possible positive and negative pairs by $\mathcal{X}^{+}$ and $\mathcal{X}^{-}$, respectively, and further define a probability distribution $p$ over $\mathcal{X}$ which, along with $f$, will yield $p^{+}$ and $p^{-}$ over $\mathcal{X}^{+}$ and $\mathcal{X}^{-}$. Similarly to the setting in \cite{hjelm2018learning}, which introduces a discriminator over pairs of samples, we are interested in $\mathcal{E}^*$ and $\mathcal{D}^*$ such that:
\begin{equation}
\begin{split}
    \mathcal{E}^*,\mathcal{D}^* & \in  \argmin_{\mathcal{E},\D} \text{  } - \E_{x^{+} \sim p^{+}} \log(\D \circ \mathcal{E}(x^{+})) \\
    & - \E_{x^{-} \sim p^{-}} \log(1-\D \circ \mathcal{E}(x^{-})),
\end{split}
\label{eq:prob_statement}
\end{equation}
and $\circ$ indicates composition so that $\D \circ \mathcal{E}(x^{+}) = \D(\mathcal{E}(x^{+}))$. Such problem is separable in the parameters of $\mathcal{E}$ and $\D$ and iterative solution strategies might include either alternate or simultaneous updates. We found the latter to converge faster in terms of wall-clock time and both approaches reach similar performance. We thus perform simultaneous updates while training.

The problem stated in (\ref{eq:prob_statement}) corresponds to finding $\mathcal{E}$ and $\D$ which will ensure that semantically close or distant samples, as defined through $f$, will preserve such properties in terms of distance in the new space, while doing so in lower dimension. We stress the fact that class labels define which samples should be close together or far apart, which means that the same underlying data can yield different pseudo-metric spaces if different semantic properties are used to define class labels. For example, if one considers that, for a given set of speech recordings, class labels are equivalent to speaker identities, recordings from the same speaker are expected to be clustered together in the embedding space, while different results can be achieved if class labels are assigned corresponding to spoken language, acoustic conditions, and so on.

\subsection{Different interpretations for $\D$}

Besides the view of $\D$ as a distance-like object defining a metric-like space $\{\mathcal{E}(\mathcal{X}),\D\}$, here we discuss some other possible interpretations of its outputs. We start by justifying the choice of the training objective defined in (\ref{eq:prob_statement}) by showing it to yield the likelihood ratio of particular trials of type I corresponding to a single enrollment example against a single test example, i.e. $T=\{x_{enroll},x_{test}\}$. In both of the next two propositions, proofs directly reuse results from the contrastive estimation and generative adversarial networks literature \cite{gutmann2010noise,goodfellow2014generative} to show $\D$ can be used for verification.

\textbf{Proposition 1.} \textit{The optimal $\D$ for any fixed $\mathcal{E}$ yields a simple transformation of the likelihood ratio stated in Eq.~\ref{eq:LR} for trials of the type $T=\{x_{enroll},x_{test}\}$}.

\textit{Proof.} We first define $p_z^{+}$ and $p_z^{-}$, which correspond to the counterparts of $p^{+}$ and $p^{-}$ induced by $\mathcal{E}$ in the embedding space. Now consider the loss $\mathcal{L}$ defined in Eq. \ref{eq:prob_statement}:
{\footnotesize{
\begin{align*}
\mathcal{L} & = - \E_{z^{+} \sim p_z^{+}} \log(\D(z^{+})) - \E_{z^{-} \sim p_z^{-}} \log(1-\D(z^{-})) \\
= & - \int\limits_{\mathcal{Z}^{+}} p_z^{+}(z^{+}) \log(\D(z^{+})) - \int\limits_{\mathcal{Z}^{-}} p_z^{-}(z^{-}) \log(1-\D(z^{-})) \\
= &- \int\limits_{\mathcal{Z}'} p_z^{+}(z') \log(\D(z')) + p_z^{-}(z') \log(1-\D(z')),\numberthis
\end{align*}}}where $\mathcal{Z}'$ corresponds to $\mathcal{Z}^{+} \cup \mathcal{Z}^{-}$ or equivalently $\mathcal{E}(\mathcal{X}^{+}) \cup \mathcal{E}(\mathcal{X}^{-})$. Since $\D(z') \in (0,1) \text{  } \forall \text{  } z' \in \mathcal{Z}'$, above integrand $p_z^{+}(z') \log(\D(z')) + p_z^{-}(z') \log(1-\D(z'))$, provided that the set from which we pick candidate solutions is rich enough, has its maximum at:
\begin{align*}
    \mathcal{D}^*(z') & = \frac{p_z^{+}(z')}{p_z^{+}(z')+p_z^{-}(z')},\\
    & = \frac{1}{1+\Big(\frac{p_z^{+}(z')}{p_z^{-}(z')}\Big)^{-1}}.\numberthis
\end{align*}
The last step above is of course only valid for $z' \in supp(p_z^{+})$. Nevertheless, $\D^*(z')$ is in any case meaningful for verification. In fact, as will be discussed in Proposition 2, the optimal encoder is the one that induces $supp(p_z^{+}) \cap supp(p_z^{-}) = \varnothing$. Considering trial $T=\{x_{enroll},x_{test}\}$, we can write the ratio $\frac{p_z^{+}(z')}{p_z^{-}(z')}$ as:
\begin{equation}
    \frac{p_z^{+}(z')}{p_z^{-}(z')} = \frac{p_z^{+}(\mathcal{E}(x_{enroll}),\mathcal{E}(x_{test}))}{p_z^{-}(\mathcal{E}(x_{enroll}),\mathcal{E}(x_{test}))} := \frac{p(T|H_0)}{p(T|H_1)}. _\square
\end{equation}

Proposition 1 indicates that the discussed setting can be used in an end-to-end fashion to yield verification decision rules against a threshold $\tau$ for trials of a specific type.

The following lemma will be necessary for the next result:

\textbf{Lemma 1.} \textit{If $supp(p_z^{+}) \cap supp(p_z^{-}) = \varnothing$, any positive threshold $0<\tau<\infty$ yields optimal decision rules for trials $T=\{x_{enroll},x_{test}\}$.}

\textit{Proof.} We prove the lemma by inspecting the decision rule under the considered assumptions in the two possible test cases: if $T$ is non-target $\implies\frac{p_z^{+}(\mathcal{E}(x_{enroll}),\mathcal{E}(x_{test}))}{p_z^{-}(\mathcal{E}(x_{enroll}),\mathcal{E}(x_{test}))} = 0 < \tau$. If $T$ is target $\implies \frac{p_z^{+}(\mathcal{E}(x_{enroll}),\mathcal{E}(x_{test}))}{p_z^{-}(\mathcal{E}(x_{enroll}),\mathcal{E}(x_{test}))} \rightarrow \infty > \tau$, completing the proof.
$_\square$

We now proceed and use the optimal discriminator into $\mathcal{L}$, which yields the following result for the optimal encoder:

\textbf{Proposition 2.} \textit{Minimizing $\mathcal{L}$ yields optimal decision rules for any positive threshold.}

\textit{Proof.} We plug $\D^*$ into $\mathcal{L}$ so that for any $z'$ we obtain:
\begin{equation}
    \begin{split}
    \mathcal{L} = & - \E_{z' \sim p_z^{+}} \log\Big(\frac{p_z^{+}(z')}{p_z^{+}(z')+p_z^{-}(z')}\Big) \\
    & - \E_{z' \sim p_z^{-}} \log \Big( \frac{p_z^{-}(z')}{p_z^{+}(z')+p_z^{-}(z')} \Big) \\
    = & - KL\Big( p_z^{+} || p_z^{+}+p_z^{-} \Big) - KL\Big( p_z^{-} || p_z^{+}+p_z^{-} \Big) \\
    = & \log 4 - 2JSD(p_z^{+} || p_z^{-}).
    \end{split}
\end{equation}

$\mathcal{L}$ is therefore minimized ($\mathcal{L}^*=0$) iff $\mathcal{E}$ yields $supp(p_z^{+}) \cap supp(p_z^{-}) = \varnothing$, which results in optimal decision rules for any positive threshold, invoking lemma 1, and assuming such encoders are available in the set one searches over.
$_\square$

We thus showed the proposed training scheme to be convenient for 2-sample tests under small sample regimes, such as in the case of verification, given that: (i) the distance model is also a discriminator which approximates the likelihood ratio of the joint distributions over positive and negative pairs\footnote{The joint distribution over negative pairs is simply the product of marginals: $p^{-}(x_i,x_j)=p(x_i)p(x_j)$.}, and the encoder will be such that it induces a high divergence across such distributions, rendering their ratio amenable to decision making even in cases where verified samples are as small as single enrollment and test examples.

On a speculative note, we provide yet another view of $\D$ by defining the kernel function $\mathcal{K}=\D$. If we assume $\mathcal{K}$ to satisfy Mercer's condition (which won't likely be the case within our setting since $\mathcal{K}$ will not be symmetric nor positive semidefinite), we can invoke Mercer's theorem and state that there is a feature map to a Hilbert space where verification can be performed through inner products. Training in the described setting could be viewed such that minimizing $\mathcal{L}$ becomes equivalent to building such a Hilbert space where classes can be distinguished by directly scoring data points one against the other. We hypothesize that constraining $\mathcal{K}$ to sets where Mercer's condition does hold might yield an effective approach for the problems we consider herein, which we intend to investigate in future work.

\subsection{Training}

We now describe the procedure we adopt to minimize $\mathcal{L}$ as well as some practical design decisions made based on empirical results. Both $\mathcal{E}$ and $\D$ are implemented as neural networks. In our experiments, $\mathcal{E}$ will be convolutional (2-d for images and 1-d for audio) while $\D$ is a stack of fully-connected layers which take as input concatenated embeddings of pairs of examples. Training is carried out with standard minibatch stochastic gradient descent with momentum. We perform simultaneous update steps for $\mathcal{E}$ and $\D$ since we observed that to be faster than alternate updates, while yielding the same performance. Standard regularization strategies such as weight decay and label smoothing \cite{szegedy2016rethinking} are also employed. We empirically found that employing an auxiliary multi-class classification loss significantly accelerates training. Since our approach requires labels to determine which pairs of examples are positive or negative, we make further use of the labels to compute such auxiliary loss, which will be indicated by $\mathcal{L}_{CE}$. To allow for computation of $\mathcal{L}_{CE}$, we project $z=\mathcal{E}(x)$ onto the simplex $\Delta^{L-1}$ using a fully-connected layer. Minimization is then performed on the sum of the two losses, i.e., we solve $\mathcal{E},\D \in \text{  }  \argmin \text{  } \mathcal{L}'=\mathcal{L}+\mathcal{L}_{CE}$, where the $CE$ subscript in $\mathcal{L}_{CE}$ indicates the multi-class cross-entropy loss.

All hyperparameters are selected with a random search over a pre-defined grid. For the particular case of the auxiliary loss $\mathcal{L}_{CE}$, besides the standard cross-entropy, we also ran experiments considering one of its so-called \textit{large margin} variations. We particularly evaluated models trained with the additive margin softmax approach \cite{wang2018additive}. The choice between the two types of auxiliary losses (standard or large margin) was a further hyperparameter and the decision was based on the random search over the two options. The grid used for hyperparameters selection along with the values chosen for each evaluation are presented in the appendix. A pseudocode describing our training procedure is presented in Algorithm \ref{alg:training}.

\begin{algorithm}[]
   \caption{Training procedure.}
   \label{alg:training}
\begin{algorithmic}
   \STATE $\mathcal{E}, \D = InitializeModels()$
   \REPEAT
   \STATE $x, y = SampleMinibatch()$
   \STATE $z = \mathcal{E}(x)$
   \STATE $z^{+} = GetAllPositivePairs(z,y)$
   \STATE $z^{-} = GetAllNegativePairs(z,y)$
   \STATE $y' = ProjectOntoSimplex(z)$
   \STATE $\mathcal{L}'=\mathcal{L}(z^{+},z^{-},\D)+\mathcal{L}_{CE}(y',y)$
   \STATE $\mathcal{E}, \D = UpdateRule(\mathcal{\mathcal{E}, \D, L}')$
   \UNTIL{Maximum number of iterations reached}
   \STATE \textbf{return} $\mathcal{E}, \D$
\end{algorithmic}
\end{algorithm}


\section{Evaluation}
\label{sec:evaluation}

\begin{table}[]
\caption{Evaluation of models trained under the proposed approach on image data.\label{tab:cifar_miniimagenet}}
\resizebox{0.47\textwidth}{!}{
\begin{tabular}{ccccc}
\hline
                                                                                               &                                             & \textit{Scoring} & \textit{EER} & \textit{1-AUC} \\ \hline
\multirow{4}{*}{\textit{Cifar-10}}                                                             & Triplet                                     & Cosine           & 3.80\%       & 0.98\%         \\ \cline{2-5} 
                                                                                               & \multirow{3}{*}{\textit{\textbf{Proposed}}} & E2E              & 3.43\%       & 0.60\%         \\
                                                                                               &                                             & Cosine           & 3.56\%       & 1.03\%         \\
                                                                                               &                                             & Cosine + E2E     & 3.42\%       & 0.80\%         \\ \hline
\multirow{4}{*}{\textit{\begin{tabular}[c]{@{}c@{}}Mini-ImageNet\\ (Validation)\end{tabular}}} & Triplet                                     & Cosine           & 28.91\%      & 21.58\%        \\ \cline{2-5} 
                                                                                               & \multirow{3}{*}{\textit{\textbf{Proposed}}} & E2E              & 28.64\%      & 21.01\%        \\
                                                                                               &                                             & Cosine           & 30.66\%      & 23.70\%        \\
                                                                                               &                                             & Cosine + E2E     & 28.49\%      & 20.90\%        \\ \hline
\multirow{4}{*}{\textit{\begin{tabular}[c]{@{}c@{}}Mini-ImageNet\\ (Test)\end{tabular}}}       & Triplet                                     & Cosine           & 29.68\%      & 22.56\%        \\ \cline{2-5} 
                                                                                               & \multirow{3}{*}{\textit{\textbf{Proposed}}} & E2E              & 29.26\%      & 22.04\%        \\
                                                                                               &                                             & Cosine           & 32.97\%      & 27.34\%        \\
                                                                                               &                                             & Cosine + E2E     & 29.32\%      & 22.24\%        \\ \hline
\end{tabular}}
\end{table}

We proceed to evaluation of the described framework and do so with four sets of experiments. In the first part of our evaluation, we run proof-of-concept experiments and make use of standard image datasets to simulate verification settings. We report results on all trials created for the test sets of Cifar-10 and Mini-ImageNet. In the former, the same 10 classes of examples appear for both train and test partitions, in what we refer to as closed-set verification. For the case of Mini-ImageNet, since that dataset was designed for few-shot learning applications, we have an open-set evaluation for verification since there are 64, 16, and 20 disjoint classes of training, validation, and test examples.

We then move on to a \textit{large scale realistic evaluation}. To this end, we make use of the recently introduced VoxCeleb corpus \cite{nagrani2017voxceleb,chung2018voxceleb2}, corresponding to audio recordings of interviews taken from youtube videos, which means there's no control over the acoustic conditions present in the data. Moreover, while most of the corpus corresponds to speech in English, other languages are also present, so that test recordings are from different speakers relative to the train data, and potentially also from different languages and acoustic environments. We specifically employ the second release of the corpus so that training data is composed of recordings from 5994 speakers while three test sets are available: (i) \textbf{VoxCeleb1 Test set}, which is made up of utterances from 40 speakers, (ii) \textbf{VoxCeleb1-E}, i.e., the complete first release of the data containing 1251 speakers, and (iii) \textbf{VoxCeleb1-H}, corresponding to a sub-set of the trials in \textbf{VoxCeleb1-E} so that non-target trials are designed to be hard to discriminate by using the meta-data to match factors such as nationality and gender of the speakers. We then report experiments performed to observe whether $\D$'s outputs present properties of actual distances, and finally check the influence of $\D$'s architecture on final performance.

Our main baselines for proof-of-concept experiments correspond to the same encoders as in the evaluation of our proposed approach, while $\D$ is dropped and replaced by the Euclidean distance. In those cases however, in order to get the most challenging baselines, we perform online selection of hard negatives. Our baselines closely follow the setting described in \cite{monteiro2019combining}. All such baselines are referred to as \textit{triplet} in the tables with results as a reference to the training loss in those cases. Unless specified, all models, baseline or otherwise, are trained from scratch, and the same computation budget is used for training and hyperparameter search for all models we trained.

Performance is assessed in terms of the difference to 1 of the area under the operating curve, indicated by 1-AUC in the tables, and also in terms of equal error rate (EER). EER indicates the operating point (i.e. threshold selection) at which the miss and false alarm rates are equal. Both metrics are better when closer to 0. We consider different strategies to score test trials. Both cosine similarity and PLDA are considered in some cases, and when the output of $\D$ is directly used as a score we then indicate it by E2E in a reference to \textit{end-to-end}\footnote{Scoring trials with cosine similarity can be also seen as end-to-end.}. We further remark that cosine similarity can also be used to score trials in our proposed setting, and we observed some performance gains when applying simple sum fusion of the two available scores. Additional implementation details are included in the appendix.

\subsection{Cifar-10 and Mini-ImageNet}

The encoder for evaluation on both Cifar-10 and Mini-ImageNet was implemented as a ResNet-18 \cite{he2016deep}. Results are reported in Table \ref{tab:cifar_miniimagenet}.

Results indicate the proposed scheme indeed yields effective inference strategies under the verification setting compared to traditional metric learning approaches, while using a more simplified training scheme since: (i) no sort of approach for harvesting hard negative pairs (e.g., \cite{schroff2015facenet,wu2017sampling}) is needed in our case, and those are usually expensive, (ii) the method does not require large batch sizes, and (iii) we employ a simple loss with no hyperparameters that have to be tuned, as opposed to margin-based triplet or contrastive losses. We further highlight that the encoders trained with the proposed approach have the possibility for trials to be further scored with cosine similarities, which yields a performance improvement in some cases when combined with $\D$'s output

\begin{figure}[]
\centering
\includegraphics[width=0.35\textwidth, trim={0 0 0 0.85cm}, clip]{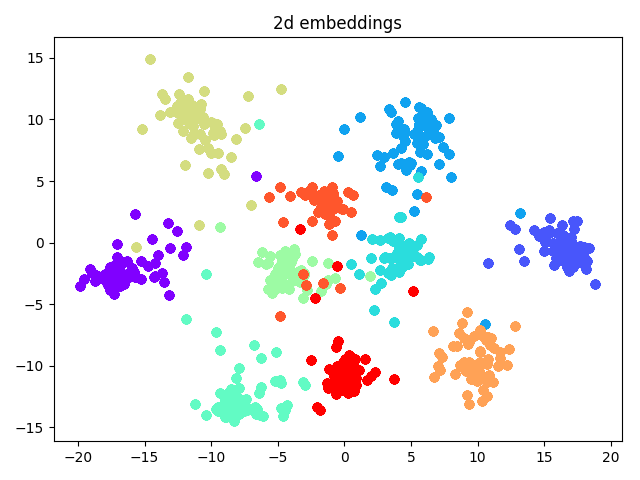}
\caption{MNIST embeddings on a 2-dimensional space. Each color represents test examples corresponding to a digit from 0 to 9.}
\label{fig:mnist}
\end{figure}

\subsection{Large-scale verification with VoxCeleb}

We now proceed and evaluate the proposed scheme in a more challenging scenario corresponding to realistic audio data for speaker verification. To do so, we implement $\mathcal{E}$ as the well-known time-delay architecture \cite{waibel1989phoneme} employed within the x-vector setting, showed to be effective in summarizing speech into speaker- and spoken language-dependent representations \cite{snyder2018x, snyder2018spoken}. The model consists of a sequence of dilated 1-dimensional convolutions across the temporal dimension, followed by a time pooling layer, which simply concatenates element-wise first- and second-order statistics over time. Statistics are finally projected into an output vector through fully-connected layers. Speech is represented as 30 mel-frequency cepstral coefficients obtained with a short-time Fourier transform using a 25ms Hamming window with 60\% overlap. All the data is downsampled to 16kHz beforehand. An energy-based voice activity detector is employed to filter out non-speech frames. We augment the data by creating noisy versions of training recordings using exactly the same approach as in \cite{snyder2018x}. Model architecture and feature extraction details are included in the appendix.

We compared our models with a set of published results as well as the results provided by the popular Kaldi recipe\footnote{\label{kaldirecipe}Kaldi recipe: \url{https://github.com/kaldi-asr/kaldi/tree/master/egs/voxceleb}}, considering scoring using cosine similarity or PLDA. For the Kaldi baseline, we found the same model as ours to yield relatively weak performance. As such, we decided to search over possible architectures in order to make it a stronger baseline. We thus report the best model we could find which has the same structure as ours, i.e., it is made up of convolutions over time followed by temporal pooling and fully-connected layers, while the convolutional stack is deeper, which makes the comparison unfair in their favor.

We further evaluated our models using PLDA by running just the part of the same Kaldi recipe corresponding to the training of that downstream classifier on top of representations obtained from our encoder. Results are reported in Table \ref{tab:voxceleb} and support our claim that the proposed framework can be directly used in an end-to-end fashion. It is further observed that it outperformed standard downstream classifiers, such as PLDA, by a significant difference while not requiring any complex training procedure, as common metric learning approaches usually do. We employ simple random selection of training pairs. Ablation results are also reported, in which case we dropped the auxiliary loss $\mathcal{L_{CE}}$ and trained the same $\mathcal{E}$ and $\D$ using the same budget in terms of number of iterations, showing that having the auxiliary loss improves performance in the considered evaluation.

\begin{table}[h]
\caption{Evaluation of models trained under the proposed approach on VoxCeleb.\label{tab:voxceleb}}
\resizebox{0.47\textwidth}{!}{
\begin{tabular}{cccc}
\hline
                            & \textit{Scoring} & \textit{Training set} & \textit{EER} \\ \hline
\textbf{VoxCeleb1 Test set} &                  &                       &              \\ \hline
\citet{nagrani2017voxceleb}              & PLDA             & VoxCeleb1             & 8.80\%       \\
\citet{Cai2018}                  & Cosine           & VoxCeleb1             & 4.40\%       \\
\citet{okabe2018attentive}                & Cosine           & VoxCeleb1             & 3.85\%       \\
\citet{hajibabaei2018unified}           & Cosine           & VoxCeleb1             & 4.30\%       \\
\citet{ravanelli2019learning}             & Cosine           & VoxCeleb1             & 5.80\%       \\
\citet{chung2018voxceleb2}                & Cosine           & VoxCeleb2             & 3.95\%       \\
\citet{xie2019utterance}                  & Cosine           & VoxCeleb2             & 3.22\%       \\
\citet{hajavi2019deep}               & Cosine           & VoxCeleb2             & 4.26\%       \\
\citet{xiang2019margin}                & Cosine           & VoxCeleb2             & 2.69\%       \\
Kaldi recipe\cref{kaldirecipe}                & PLDA             & VoxCeleb2             & 2.51\%       \\
\textit{\textbf{Proposed}}  & Cosine           & VoxCeleb2             & 4.97\%       \\
\textit{\textbf{Proposed}}  & E2E              & VoxCeleb2             & 2.51\%       \\
\textit{\textbf{Proposed}}  & Cosine + E2E     & VoxCeleb2             & 2.51\%       \\
\textit{\textbf{Proposed}}  & PLDA             & VoxCeleb2             & 3.75\%       \\ \hline
\textit{\textbf{Ablation} $(-\mathcal{L}_{CE})$}  & E2E             & VoxCeleb2             & 3.44\%       \\ \hline
\textbf{VoxCeleb1-E}        &                  &                       &              \\ \hline
\citet{chung2018voxceleb2}                & Cosine           & VoxCeleb2             & 4.42\%       \\
\citet{xie2019utterance}                  & Cosine           & VoxCeleb2             & 3.13\%       \\
\citet{xiang2019margin}                & Cosine           & VoxCeleb2             & 2.76\%       \\
Kaldi recipe\cref{kaldirecipe}                & PLDA             & VoxCeleb2             & 2.60\%       \\
\textit{\textbf{Proposed}}  & Cosine           & VoxCeleb2             & 4.77\%       \\
\textit{\textbf{Proposed}}  & E2E              & VoxCeleb2             & 2.57\%       \\
\textit{\textbf{Proposed}}  & Cosine + E2E     & VoxCeleb2             & 2.53\%       \\
\textit{\textbf{Proposed}}  & PLDA             & VoxCeleb2             & 3.61\%       \\ \hline
\textit{\textbf{Ablation} $(-\mathcal{L}_{CE})$}  & E2E             & VoxCeleb2             & 3.70\%       \\ \hline
\textbf{VoxCeleb1-H}        &                  &                       &              \\ \hline
\citet{chung2018voxceleb2}                & Cosine           & VoxCeleb2             & 7.33\%       \\
\citet{xie2019utterance}                  & Cosine           & VoxCeleb2             & 5.06\%       \\
\citet{xiang2019margin}                & Cosine           & VoxCeleb2             & 4.73\%       \\
Kaldi recipe\cref{kaldirecipe}                & PLDA             & VoxCeleb2             & 4.62\%       \\
\textit{\textbf{Proposed}}  & Cosine           & VoxCeleb2             & 8.61\%       \\
\textit{\textbf{Proposed}}  & E2E              & VoxCeleb2             & 4.73\%       \\
\textit{\textbf{Proposed}}  & Cosine + E2E     & VoxCeleb2             & 4.69\%       \\
\textit{\textbf{Proposed}}  & PLDA             & VoxCeleb2             & 5.98\%       \\ \hline
\textit{\textbf{Ablation} $(-\mathcal{L}_{CE})$}  & E2E             & VoxCeleb2             & 7.76\%       \\ \hline
\end{tabular}}
\end{table}

\begin{figure*}[]
	\centering
	\subfloat[fig:met_cifar][Distance to itself - Cifar-10.]{\includegraphics[width=0.3\textwidth, trim={0 6.1cm 0 6.1cm}, clip]{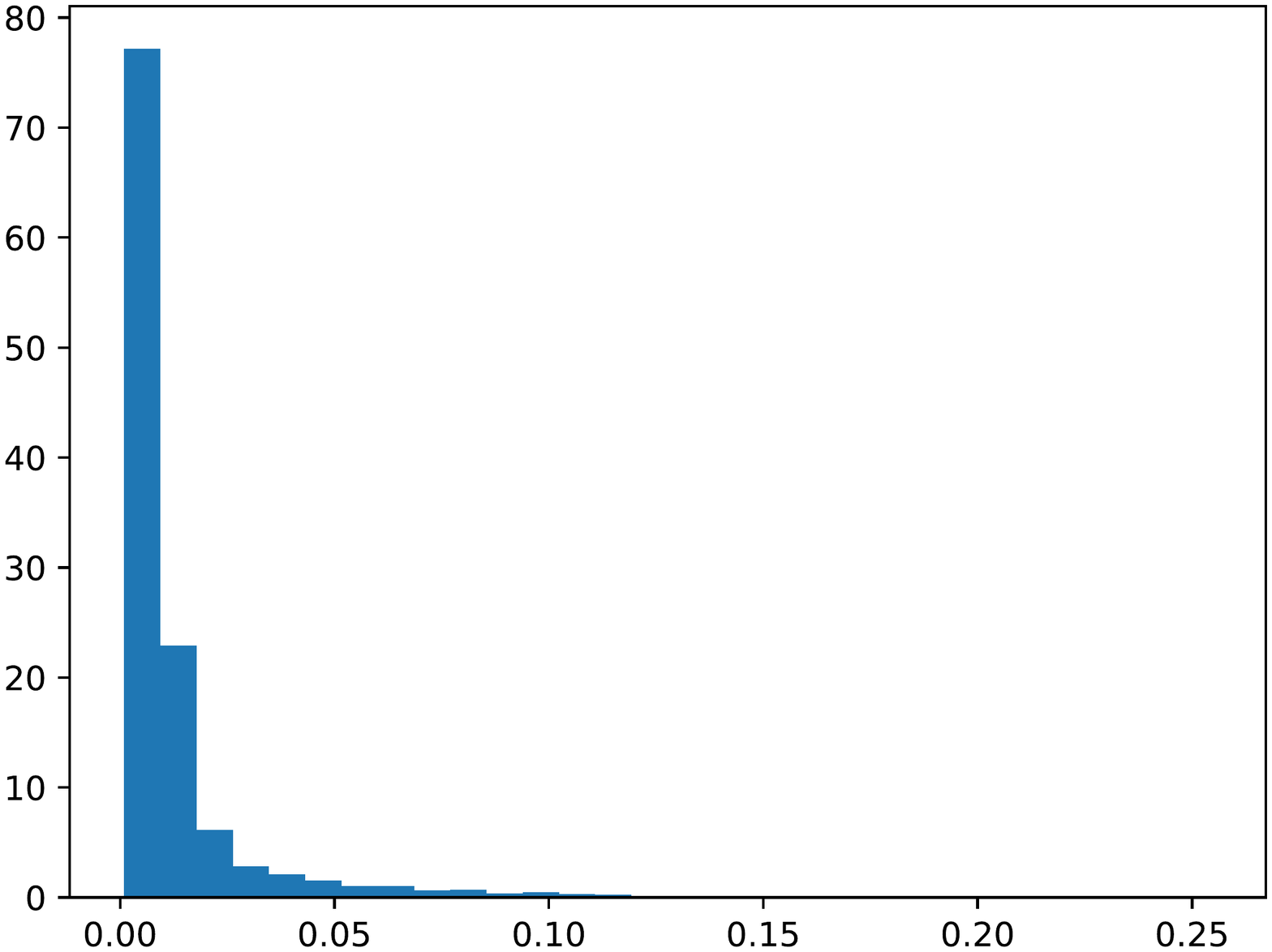}}
	\quad
	\subfloat[fig:sym_cifar][Symmetry -  Cifar-10.]{\includegraphics[width=0.3\textwidth, trim={0 6.1cm 0 6.1cm}, clip]{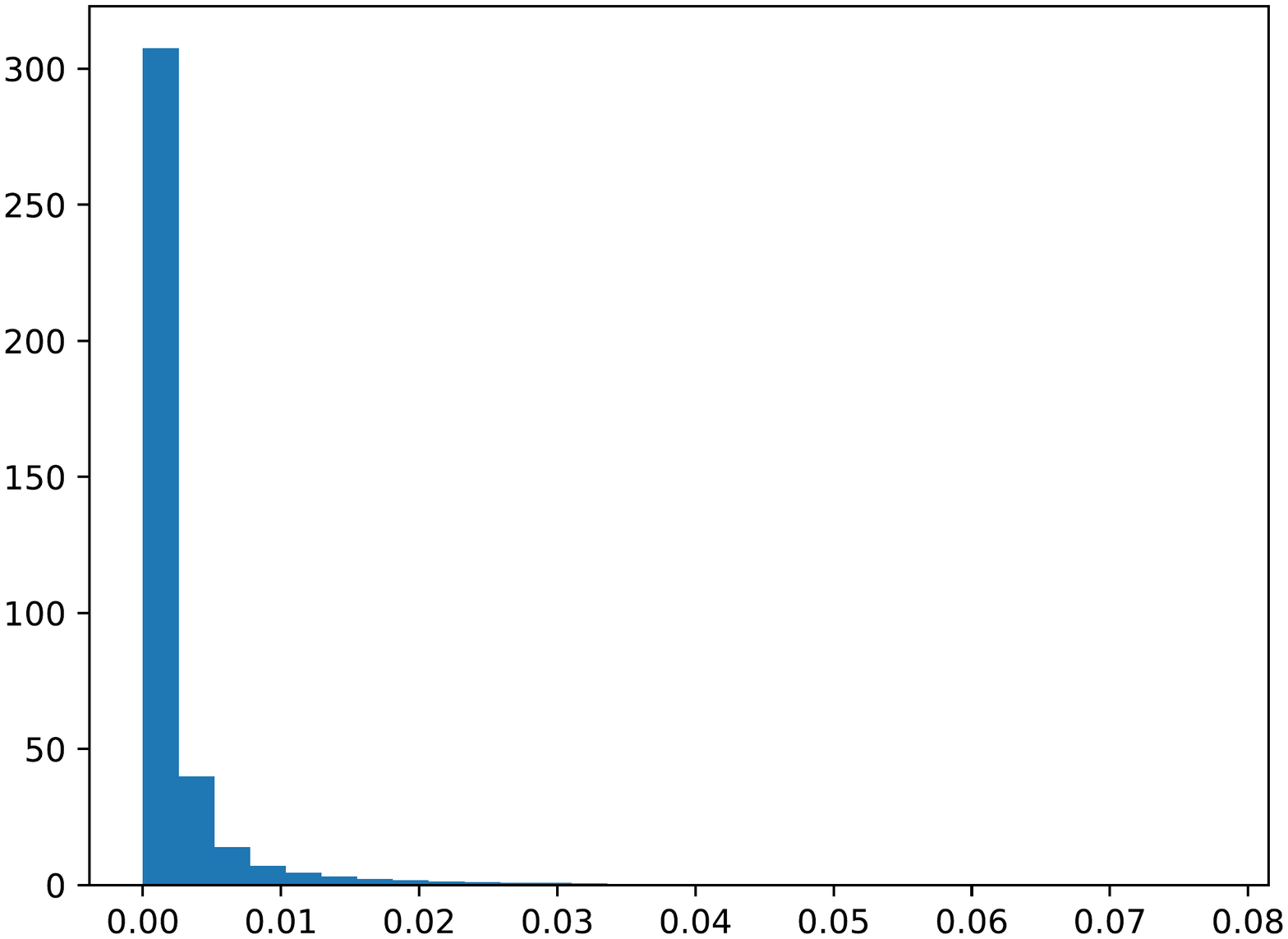}}
    \quad
	\subfloat[fig:triang_cifar][Triangle inequality - Cifar-10.]{\includegraphics[width=0.3\textwidth, trim={0 6.1cm 0 6.1cm}, clip]{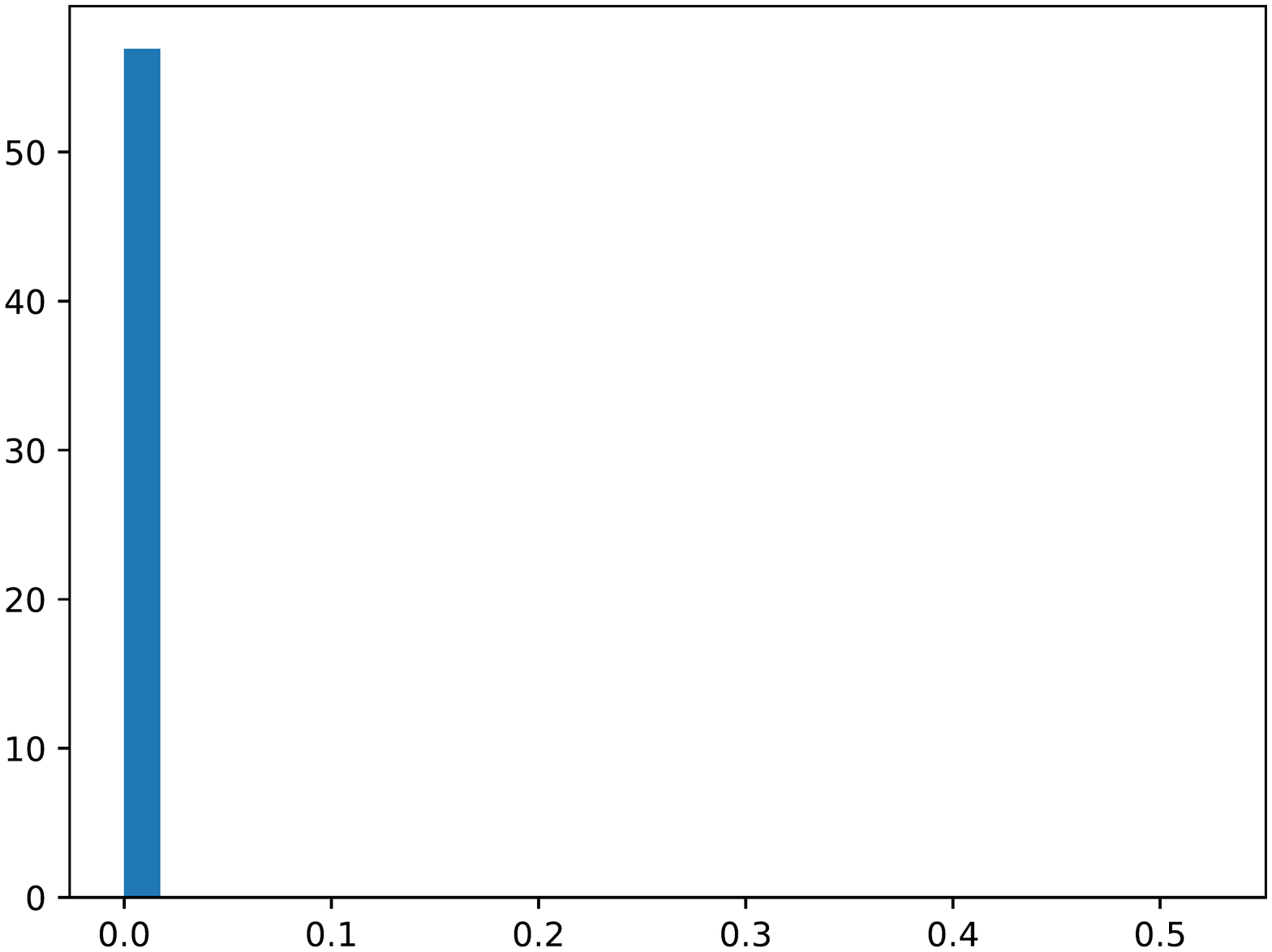}}
	\quad
	\subfloat[fig:met_vox][Distance to itself - VoxCeleb.]{\includegraphics[width=0.3\textwidth, trim={0 6.1cm 0 6.1cm}, clip]{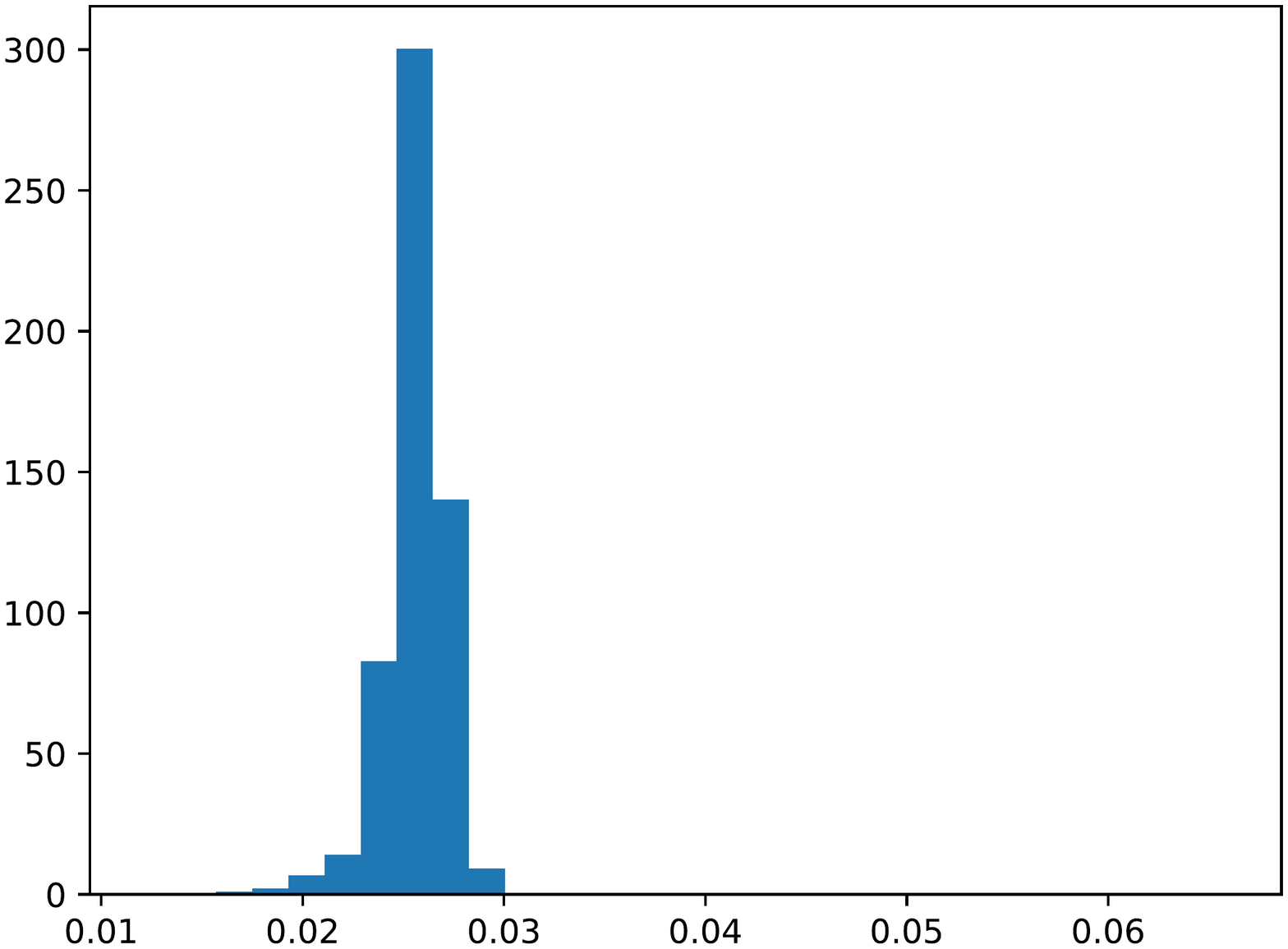}}
	\quad
	\subfloat[fig:sym_vox][Symmetry - VoxCeleb.]{\includegraphics[width=0.3\textwidth, trim={0 6.1cm 0 6.1cm}, clip]{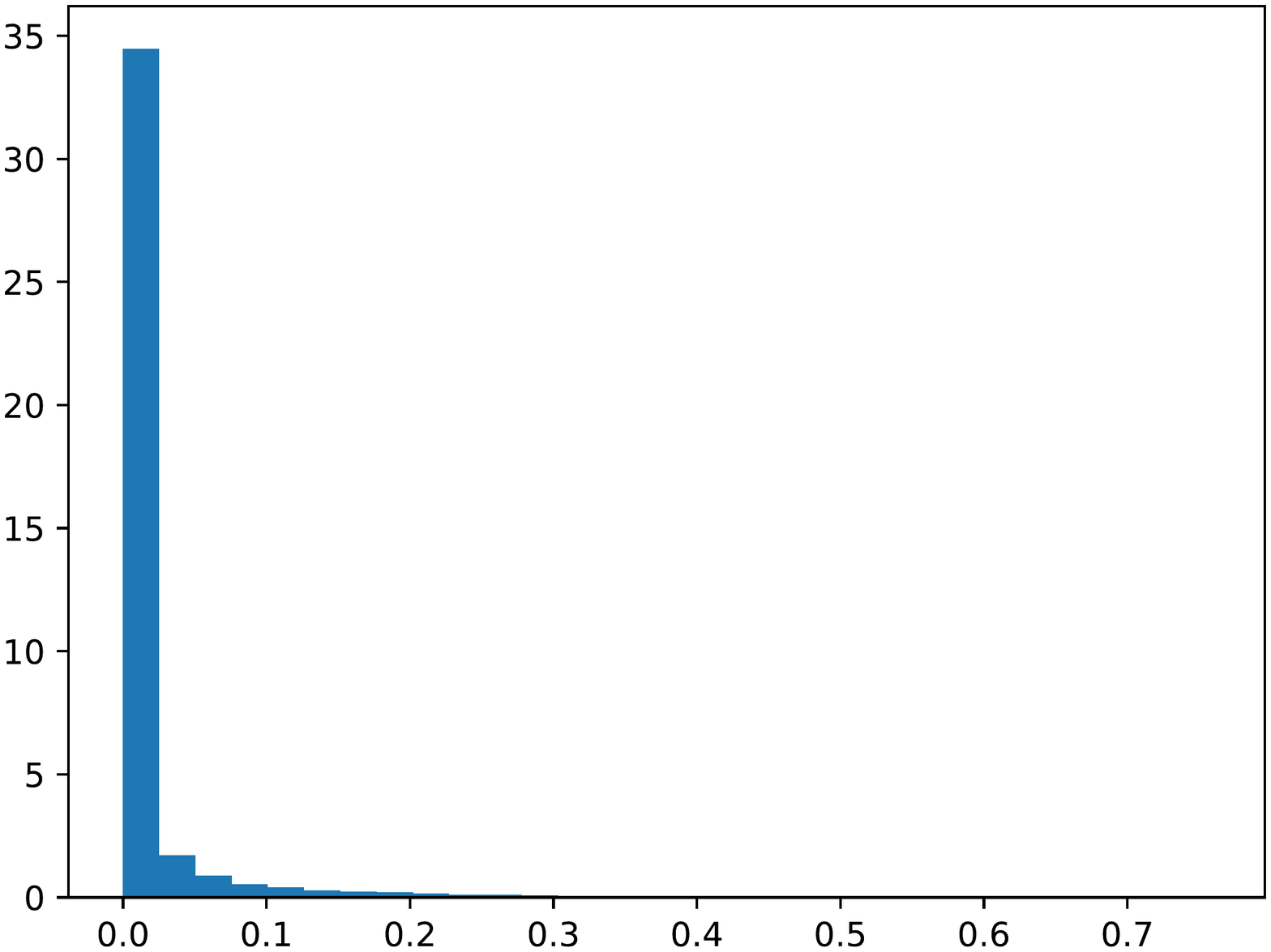}}
	\quad
	\subfloat[fig:triang_vox][Triangle inequality - VoxCeleb.]{\includegraphics[width=0.3\textwidth, trim={0 6.1cm 0 6.1cm}, clip]{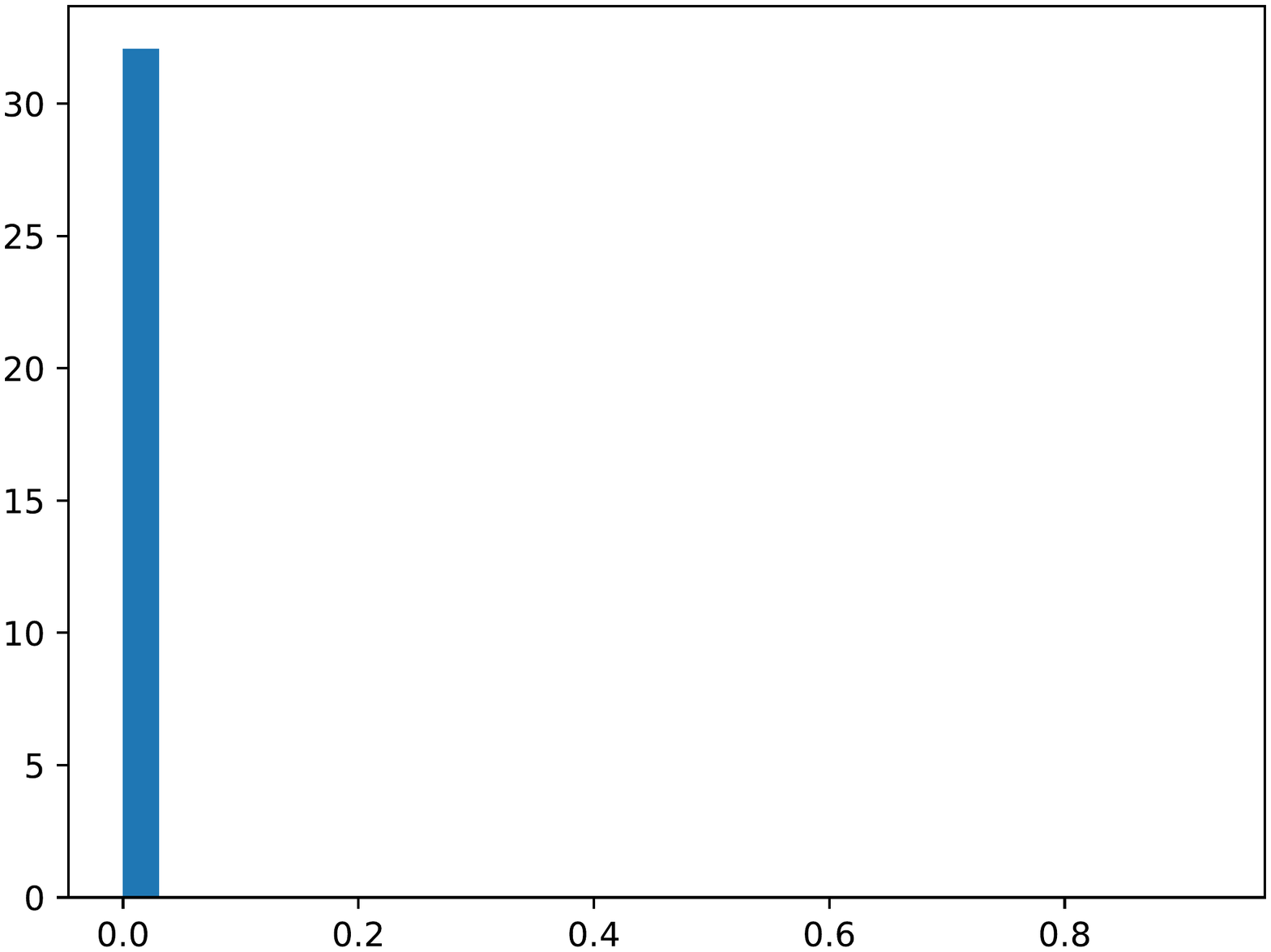}}
	\caption{Evaluation of properties given by outputs of $\D'=1-\D$. } 
	\label{fig:properties}
\end{figure*}

\subsection{Checking for distance properties in $\D$}

We now empirically evaluate how $\D$ behaves in terms of properties of distances or metrics such as symmetry, for instance. We start by plotting embeddings from $\mathcal{E}$ and do so by training an encoder on MNIST under the proposed setting (without the auxiliary loss $\mathcal{L}_{CE}$ in this case) so that its outputs are given by $z \in \R^2$. We then plot the embeddings of the complete MNIST's test set on Fig.~\ref{fig:mnist}, where the raw embeddings in $\R^2$ are directly displayed in the plot. Interestingly, classes are reasonably clustered in the Euclidean space even if such behavior was never enforced during training. We proceed and directly check for distance properties in $\D'=1-\D$. For the test set of Cifar-10 as well as for \textbf{VoxCeleb1 Test set}, we plot histograms of (i) the distance to itself for all test examples, (ii) a symmetry measure given by the absolute difference of the outputs of $\D'$ measured in the two directions for all possible test pairs, and (iii) a measure of how much $\D'$ satisfies the triangle inequality, which we do by measuring $\max[\D'(b,c)-(\D'(a,b)+\D'(a,c)), 0]$ for a random sample taken from all possible triplets of examples $\{a,b,c\}$. Proper metrics should have all such quantities equal 0. In Figures \ref{fig:properties}-a to \ref{fig:properties}-f, it can be seen that once more, even if any particular behavior is enforced over $\D$ at its training phase, resulting models approximately behave as proper metrics. We thus hypothesize the relatively easier training observed in our setting, in the sense that it works without complicated schemes for selection of negative pairs, is due to the not so constrained distances induced by $\D$.

\subsection{Varying the depth of $\D$ for verification on ImageNet}

\begin{figure*}[]
	\centering
	\subfloat[fig:imagenet_eer][EER.]{\includegraphics[width=0.489\textwidth, trim={0.65cm 0.65cm 0.9cm 0.7cm}, clip]{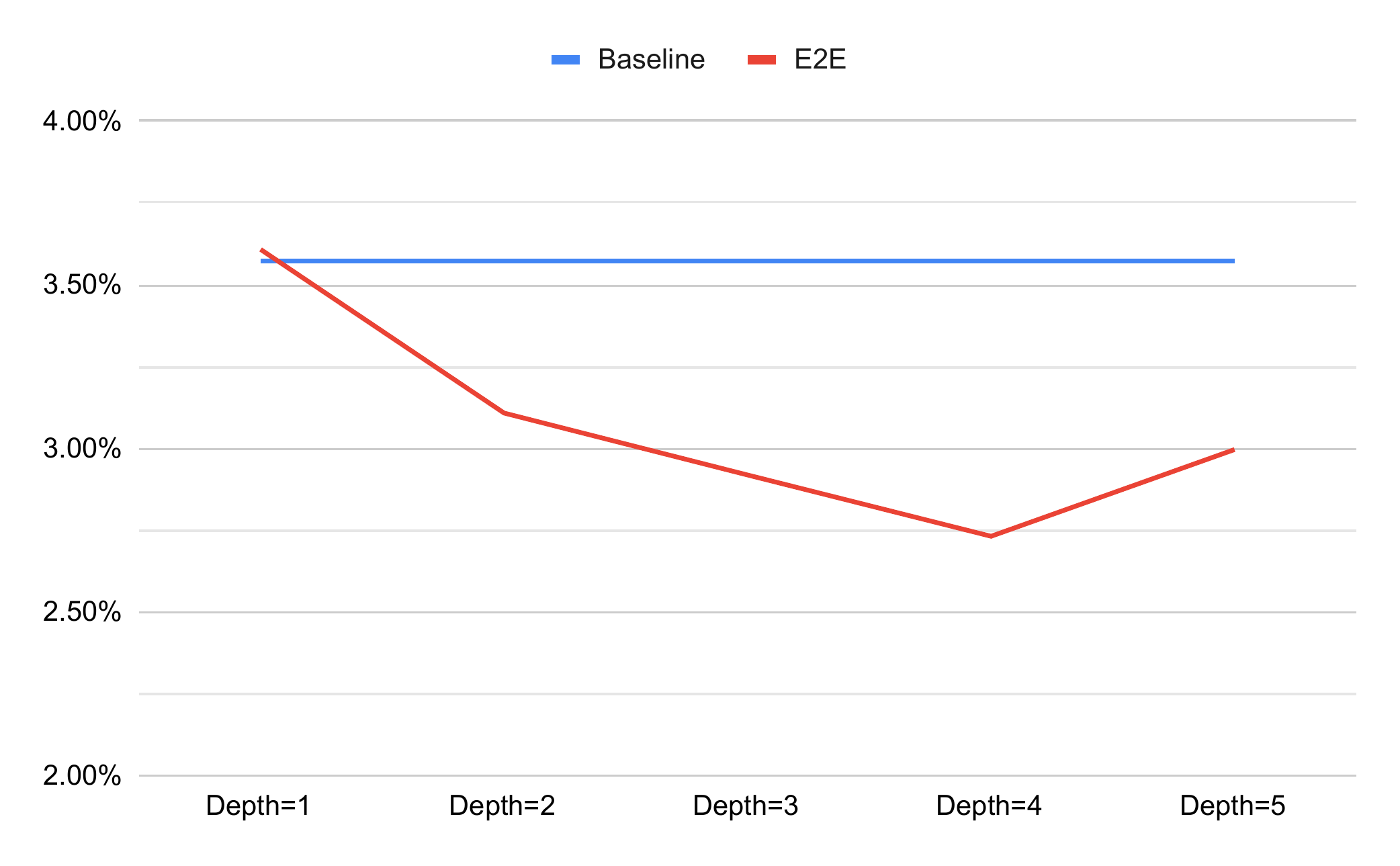}}
	\quad
	\subfloat[fig:imagenet_auc][1-AUC.]{\includegraphics[width=0.489\textwidth, trim={0.65cm 0.65cm 0.9cm 0.7cm}, clip]{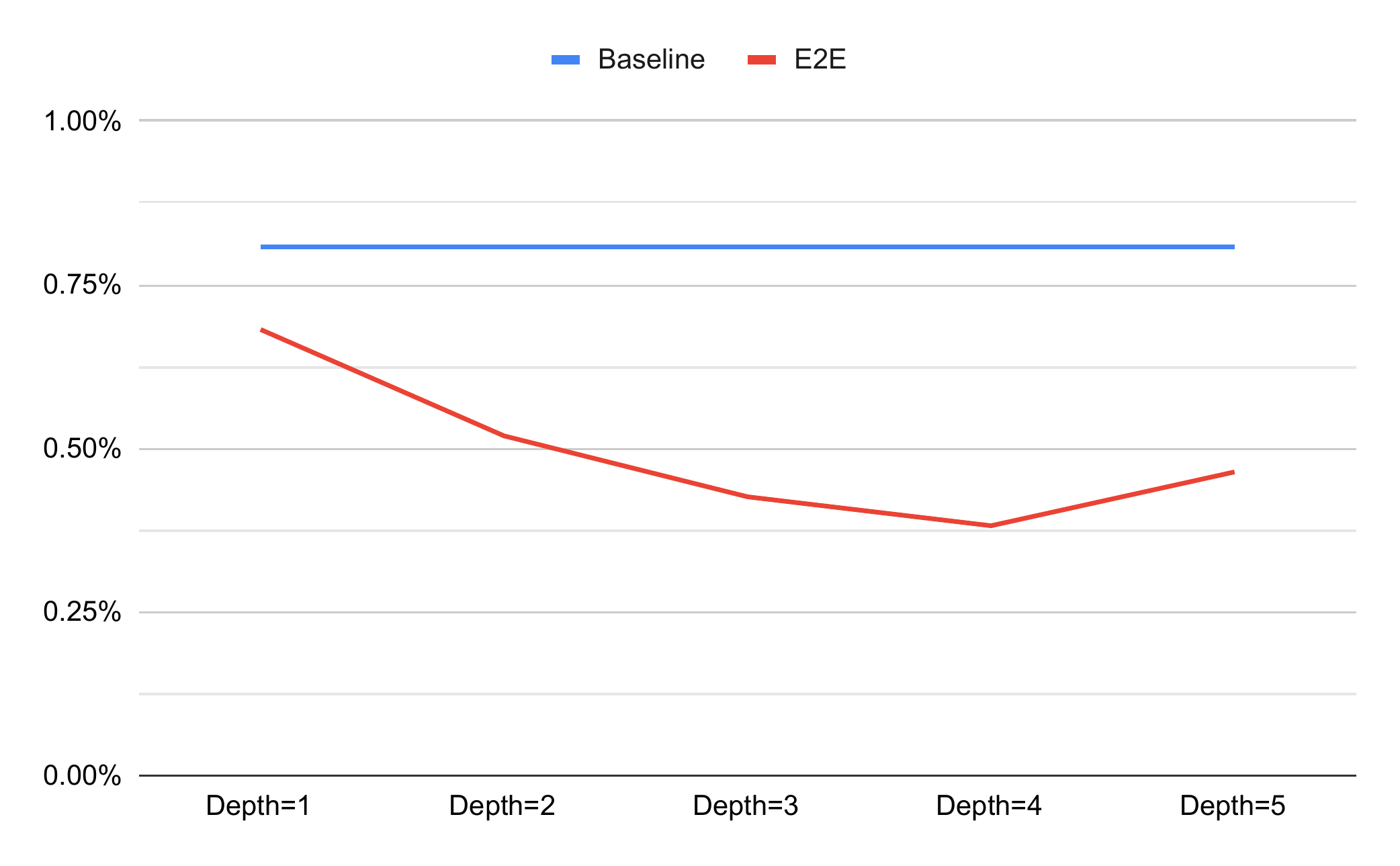}}
	\caption{Verification on all trials created by pairing all test examples of ImageNet. Results indicate that defining the architecture of the distance model is not difficult in practice given that models of varying depths yield a relatively small performance range.}
	\label{fig:imagenet}
\end{figure*}

We performed closed-set verification on the full ImageNet with distance models of increasing depths (1 to 5) to verify whether our setting is stable with respect to some of the introduced hyperparameters. With this experiment, we specifically intend to assess how difficult it would be in practice to find a good architecture for the distance model. Our models are compared against encoders with the same architecture, but trained using a standard metric learning approach, i.e the same training scheme as that employed for baselines reported in Table \ref{tab:cifar_miniimagenet}.

For this case, the encoder $\mathcal{E}$ is implemented as the convolutional stack of a ResNet-50 followed by a fully-connected layer used to project the output representations to the desired dimensionality, and we employ an embedding dimension of 128 across all reported models. $\D$ is once more implemented as a stack of fully-connected layers in which case we set the sizes of all hidden layers to 256. Training is performed such that the parameters of the convolutional portion of $\mathcal{E}$ are initialized from a pretrained model for multi-class classification on ImageNet, and this approach is used for both our models as well as the baseline. We then perform stochastic gradient descent on the combined loss discussed in Section \ref{sec:method} using the standard multi-class cross entropy as auxiliary loss. Moreover, given the large number of classes in ImageNet compared to commonly used batch sizes, in order to be able to always find positive pairs throughout training, minibatches are constructed using the same strategy as that employed for experiments with \textit{VoxCeleb}, i.e. we ensure at least 5 examples per class appear in each minibatch. The learning rate is set to 0.001 and is reduced by a factor of 0.1 every 10 epochs. Training is carried out for 50 epochs. Evaluation is performed over trials obtained from building all possible pairs of examples from the test partition of ImageNet. Results are reported in Figures \ref{fig:imagenet}-a and \ref{fig:imagenet}-b in terms of EER and the area over the operating curve (1-AUC), respectively. Scoring for the case of baseline encoders is performed with cosine similarity between encoded examples from test trials. While standard metric learning encoders make for strong baselines, all evaluated distance models are able to perform on pair (depth=1) or better than (depth$>$1) such models.

The results discussed herein provide empirical evidence for the claim that tuning the hyperparameters we introduced in comparison to previous settings, i.e. the architecture of the distance model, is not so challenging in that we achieve reasonably stable performance for verification on ImageNet when varying the depth of the distance model. Yet another empirical finding supporting that claim consists of the fact that similar architectures of the distance model were found to work well across all the datasets/domains we evaluated on. We specifically found that distance models with 3 or 4 hidden layers with 256 units each work well across datasets, which we believe might be a reasonable starting point for extending the approach we discussed to other datasets.

\section{Conclusion}
\label{sec:discussion}

We introduced an end-to-end setting particularly tailored to perform small sample 2-sample tests and compare data pairs to determine whether they belong to the same class. Several interpretations of such framework are provided, including joint encoder and distance metric learning, as well as contrastive estimation over data pairs. We used contrastive estimation results to show the solutions of the posed problem yield optimal decision rules under verification settings, resulting in correct decisions for any choice of threshold. In terms of practical contributions, the proposed method simplifies both the training under the metric learning framework, as it does not require any scheme to select negative pairs of examples, and also simplifies verification pipelines, which are usually made up of several individual components, each one contributing specific challenges at training and testing phases. Our models can be used in an end-to-end fashion by using $\D$'s outputs to score test trials yielding strong performance even in large scale and realistic open-set conditions where test classes are different from those seen at train time\footnote{Code for reproducing our experiments can be found at: \url{https://github.com/joaomonteirof/e2e_verification}}. The proposed approach can be extended to any setting relying on distances to do inference such as image retrieval, prototypical networks \cite{snell2017prototypical}, and clustering. Similarly to extensions of GANs \cite{nowozin2016f,arjovsky2017wasserstein}, variations of our approach where $\mathcal{E}$ maximizes other types of divergences instead of Jensen-Shannon's might also be a relevant future research direction, requiring corresponding decision rules to be defined.

\section*{Acknowledgements}
\label{sec:acknowledgements}
Part of the authors wish to acknowledge funding from the Natural Sciences and Engineering Research Council of Canada (NSERC) through grant RGPIN-2019-05381. The first author was partially funded by the \textit{Bourse du CRIM pour \'Etudes Sup\'erieures}.

\bibliography{bibliography.bib}
\bibliographystyle{icml2020.bst}

\appendix

\clearpage

\onecolumn

\section{Implementation details}

\subsection{$\D$ architecture}

$\D$ is implemented as a stack of fully-connected layers with LeakyReLU activations. Dropout is further used in between the last hidden and the output layer. The number and size of hidden layers as well as the dropout probability were tuned for each experiment.

\subsection{Cifar-10 and Mini-ImageNet}

\subsubsection{Hyperparamters}

The grid used on the hyperparameter search for each hyperparameter is presented next. A budget of 100 runs was considered and each model was trained for 600 epochs. Hyperparameters yielding the best EER on the validation data for our proposed approach and the triplet baseline are represented by $^*$ and $^{\dagger}$, respectively. In all experiments, the minibatch size was set to 64 and 128 for Cifar-10 and Mini-ImageNet, respectively. A reduce-on-plateau schedule for the learning rate was employed, while its patience was a further hyperparameter included in the search.

\paragraph*{Cifar-10:}

\begin{itemize}
    \item Learning rate: $\{0.5, 0.1, 0.01^{*, \dagger}, 0.001\}$
    \item Weight decay: $\{0.01, 0.001^{*}, 0.0001^{\dagger}, 0.00001\}$
    \item Momentum: $\{0.1, 0.5, 0.9^{*, \dagger}\}$
    \item Label smoothing: $\{0.01, 0.1, 0.2^{*, \dagger}\}$
    \item Patience: $\{1, 5, 10^{*, \dagger}, 20\}$
    \item Number of $\D$ hidden layers: $\{2, 3^{*}, 4, 5 \}$
    \item Size of $\D$ hidden layers: $\{128, 256, 350^{*}, 512 \}$
    \item $\D$ dropout probability: $\{0.01, 0.1, 0.2^{*}\}$
    \item Type of auxiliary loss: $\{$Standard cross-entropy, Additive margin$^{*, \dagger}\}$
\end{itemize}

\paragraph*{Mini-Imagenet:}

\begin{itemize}
    \item Learning rate: $\{0.5, 0.1^{\dagger}, 0.01^{*}, 0.001\}$
    \item Weight decay: $\{0.01, 0.001^{*}, 0.0001^{\dagger}, 0.00001\}$
    \item Momentum: $\{0.1, 0.5, 0.9^{*, \dagger}\}$
    \item Label smoothing: $\{0.01, 0.1^{*}, 0.2^{\dagger}\}$
    \item Patience: $\{1, 5, 10^{*, \dagger}, 20\}$
    \item Number of $\D$ hidden layers: $\{2, 3^{*}, 4, 5 \}$
    \item Size of $\D$ hidden layers: $\{128, 256, 350^{*}, 512 \}$
    \item $\D$ dropout probability: $\{0.01, 0.1^{*}, 0.2\}$
    \item Type of auxiliary loss: $\{$Standard cross-entropy, Additive margin$^{*, \dagger}\}$
\end{itemize}

\subsection{Voxceleb}

\subsubsection{Encoder architecture}

We implement $\mathcal{E}$ as the well-known TDNN architecture employed within the x-vector setting \cite{snyder2018x}, which consists of a sequence of dilated 1-dimensional convolutions across the temporal dimension, followed by a time pooling layer, which simply concatenates element-wise first- and second-order statistics over time. Concatenated statistics are finally projected into an output vector through two fully-connected layers. Pre-activation batch normalization is performed after each convolution and fully-connected layer. A summary of the employed architecture is shown in Table~\ref{tab:TDNN_arch}.

\begin{table}[h]
\centering
\caption{Employed TDNN encoder. $T$ indicates the duration of features in number of frames.}
\begin{tabular}{ccc}
\hline
\textit{\textbf{Layer}}      & \textit{\textbf{Input Dimension}} & \textit{\textbf{Output dimension}} \\ \hline
\textit{Conv1d+ReLU}         & 30 $\times$ T                     & 512 $\times$ T                     \\
\textit{Conv1d+ReLU}         & 512 $\times$ T                    & 512 $\times$ T                     \\
\textit{Conv1d+ReLU}         & 512 $\times$ T                    & 512 $\times$ T                     \\
\textit{Conv1d+ReLU}         & 512 $\times$ T                    & 512 $\times$ T                     \\
\textit{Conv1d+ReLU}         & 512 $\times$ T                    & 1500 $\times$ T                    \\
\textit{Statistical Pooling} & 1500 $\times$ T                   & 3000                               \\
\textit{Linear+ReLU}         & 3000 $\times$ T                   & 512                               \\
\textit{Linear+ReLU}         & 512                              & $d$                                \\ \hline
\end{tabular}
\label{tab:TDNN_arch}
\end{table}

\subsubsection{Data augmentation and feature extraction}

We augment the training data by simulating diverse acoustic conditions using supplementary noisy speech, as done in \cite{snyder2018x}. More specifically, we corrupt the original samples by adding reverberation (reverberation time varies from 0.25s - 0.75s) and background noise such as music (signal-to-noise ratio, SNR, within 5-15dB), and babble (SNR varies from 10dB to 20dB). Noise signals were selected from the MUSAN corpus \cite{musan2015} and the room impulse responses samples from \cite{ko2017study} were used to simulate reverberation. All the audio pre-processing steps including feature extraction, degradation with noise as well as silence frames removal was performed with the Kaldi toolkit \cite{povey2011kaldi} and are openly available as the first step of the recipe in \url{https://github.com/kaldi-asr/kaldi/tree/master/egs/voxceleb}. The corpora used for augmentation are also openly available at \url{https://www.openslr.org/}.

In order to deal with recordings of varying duration within a minibatch, we pad all recordings to a maximum duration set in advance. We do so by repeating the signal up until it reaches the maximum duration or taking a random continuous chunk with the maximum duration for the case of long utterances.

\subsubsection{Minibatch construction}

Given the large number of classes in the VoxCeleb case (corresponding to the number of speakers, i.e., 5994), we need to ensure several examples belonging to the same speaker exist in a minibatch to allow for positive pairs to exist. We thus create a list of sets of five recordings belonging to the same speaker, and such sets are randomly selected at training time. Minibatches are constructed through sequentially picking examples from the list, and the list is recreated once all elements are sampled. Such approach provides minibatches of size $N_e = S \cdot R$, where $R$ and $S$ correspond to the number of speakers per minibatch and recordings per speaker, respectively. While $R$ is set to 5, $S$ is set to 24, which gives an effective minibatch size  of $N_e = 120$.

\subsubsection{Hyperparamters}

Training was carried out with a linear learning rate warm-up, employed during the first iterations, and the same exponential decay as in \cite{vaswani2017attention} is employed after that. A budget of 40 runs was considered and each model was trained for a budget of 600k iterations. The best set of hyperparameters, as assessed in terms of EER measured over a random set of trials created from \textbf{VoxCeleb1-E}, was then used to train a model from scratch for a total of 2M iterations. We report the results obtained by the best model within the 2M iterations in terms of the same metric used during the hyperparameter search. Selected values are indicated by $^{*}$.

The grid used for the hyperparameter search is presented next. In all experiments, the minibatch size was set to 24, which, given the sampling strategy employed in this case, yields an effective batch size of 120. We further employed gradient clipping and searched over possible clipping thresholds.

\begin{itemize}
    \item Base learning rate: $\{2.0, 1.5^{*}, 1.0, 0.5, 0.1\}$
    \item Weight decay: $\{0.001^{*}, 0.0001, 0.00001\}$
    \item Momentum: $\{0.7, 0.85, 0.95^{*}\}$
    \item Label smoothing: $\{0.0, 0.1^{*}, 0.2]\}$
    \item Embedding size $d$: $\{128, 256^{*}, 512 \}$
    \item Maximum duration (in number of frames): $\{300, 500, 800^{*}\}$
    \item Gradient clipping threshold: $\{10^{*}, 20, 50\}$
    \item Number of $\D$ hidden layers: $\{1, 2, 3, 4^{*} \}$
    \item Size of $\D$ hidden layers: $\{128, 256^{*}, 350, 512 \}$
    \item $\D$ dropout probability: $\{0.01, 0.1^{*}, 0.2\}$
    \item Type of auxiliary loss: $\{$Standard cross-entropy, Additive margin$^{*}\}$
\end{itemize}

\clearpage

\section{Large scale speaker verification under domain shift}

In this experiment, we evaluate the performance of the proposed setting when test data significantly differs from training examples. To do so, we employ the data introduced for one of the tasks of the 2018 edition of the NIST Speaker Recognition Evaluation (SRE)\footnote{\url{https://www.nist.gov/system/files/documents/2018/08/17/sre18_eval_plan_2018-05-31_v6.pdf}}. We specifically consider the CTS task so that test data corresponds to spontaneous conversational telephone speech spoken in Tunisian Arabic, while the bulk of the train data is spoken in English. Besides the language mismatch, variations due to different codecs are further observed (PSTN vs. PSTN and VOIP).

The main training dataset (English) is built by combining the data from \textit{NIST SRE}s from 2004 to 2010, \textit{Mixer} 6, as well as \textit{Switchboard}-2, phases 1, 2, and 3, and the first release of \textit{VoxCeleb}, yielding a total of approximately 14000 speakers. Audio representations correspond to 23 MFCCs obtained using a short-time Fourier transform with a 25ms Hamming window and 60\% overlap. The audio data is downsampled to 8kHz. Further pre-processing steps are the same as those performed for experiments with VoxCeleb as reported in Section \ref{sec:evaluation}, i.e. an energy-based voice activity detector is followed by data augmentation performed via distorting original samples adding reverberation and background noise.

\textbf{Baseline:} For performance reference, we trained the well-known x-vector setting \cite{snyder2018x} using its Kaldi recipe\footnote{\url{https://github.com/kaldi-asr/kaldi/tree/master/egs/sre16/v1/local/nnet3/xvector}}. In that case, PLDA is employed for scoring test trials. The same training data used to train our systems is employed in this case as well. The recipe performs the following steps: \textbf{i}-training of a TDNN (same architecture as in our case) as a multi-class classifier over the set of training speakers using the same training data utilized to train our proposed model; \textbf{ii}-preparation of PLDA's training data, in which case the \textit{SRE} partition of the training set is encoded using the second to last layer of the TDNN, embeddings are length-normalized and mean-centered using the average of an unlabelled sample from the target domain and finally have their dimensionality reduced using Linear Discriminant Analysis; \textbf{iii}-training of PLDA; \textbf{iv}-scoring of test trials. In addition to that, in order to cope with the described domain shift, the model adaptation scheme introduced in \cite{garcia2014unsupervised} is also utilized for PLDA so that a second PLDA model is trained on top of target data. The final downstream classifier is then obtained by averaging the parameters of the original and target domain models. Both results obtained with and without the described scheme are reported in Table \ref{tab:sre}. 

\begin{table}[h]
\centering
\caption{Evaluation of models under domain shift. Target data corresponds to speech spoken in Arabic. Fine-tuning on datasets including target data yields an improvement in verification performance.\label{tab:sre}}
\begin{tabular}{cccc}
\hline
\textit{}                                   & \textit{Training domain} & \textit{Scoring} & \textit{EER} \\ \hline
\multirow{2}{*}{\citet{snyder2018x}} & English                  & PLDA             & 11.30\%      \\ \cdashline{2-4} 
                                            & English+Arabic                   & Adapted PLDA     & 9.44\%       \\ \hline
\multirow{2}{*}{\textit{\textbf{Proposed}}} & English                  & E2E              & 13.61\%      \\ \cdashline{2-4} 
                                            & Multi-language           & E2E              & 8.43\%       \\ \hline
\end{tabular}
\end{table}

For the case of the proposed approach, training is carried out using the training data described above corresponding to speech spoken in English. We reuse the setting found to work well on the experiments reported in Section \ref{sec:evaluation} with the VoxCeleb corpus including all hyperparameters, architecture, data sampling and minibatch construction strategies, and computational budget. We additionally build a multi-language training set including data corresponding to the target domain so that we can fine-tune our model. The complementary training data corresponds to the data introduced for the 2012 (English) and 2016 (Cantonese+Tagalog) editions of NIST SRE as well as the development partition of NIST SRE 2018 which corresponds to the target domain of evaluation data (Arabic). This is done so as to increase the amount of data within the complementary partition and avoid overfitting to the small amount of target data. The combination of such data sources yields approximately 800 speakers. We thus train our models on the large out-of-domain dataset and fine-tune the resulting model in the multi-language complementary data.

Results in terms of equal error rate are presented in Table \ref{tab:sre}. While our model appears to be more domain dependent when compared to PLDA as indicated by results where only out-of-domain data is employed, it significantly improves once a relatively small amount of target domain data is provided. We stress the fact that the proposed setting dramatically simplifies verification pipelines and completely removes practical issues such as those related to processing steps prior to training of the downstream classifier.

\clearpage

\section{Image retrieval and clustering}

Even though the setting we introduced is tailored to verification and we only have guarantees for that case, we further verified its performance on other tasks and once again concluded it reaches competitive performance while using a much simpler and general training/testing workflow. Such extra evaluation is performed using well-known image retrieval benchmarks: Caltech birds\footnote{\url{http://www.vision.caltech.edu/visipedia/CUB-200-2011.html}} (CUB), CARS196\footnote{\url{https://ai.stanford.edu/~jkrause/cars/car_dataset.html}}, and Stanford online products\footnote{\url{https://cvgl.stanford.edu/projects/lifted_struct/}} (SOP).

\begin{figure}[h]
	\centering
	\subfloat[fig:retrieval_cub][R@k - CUB.]{\includegraphics[width=0.48\textwidth]{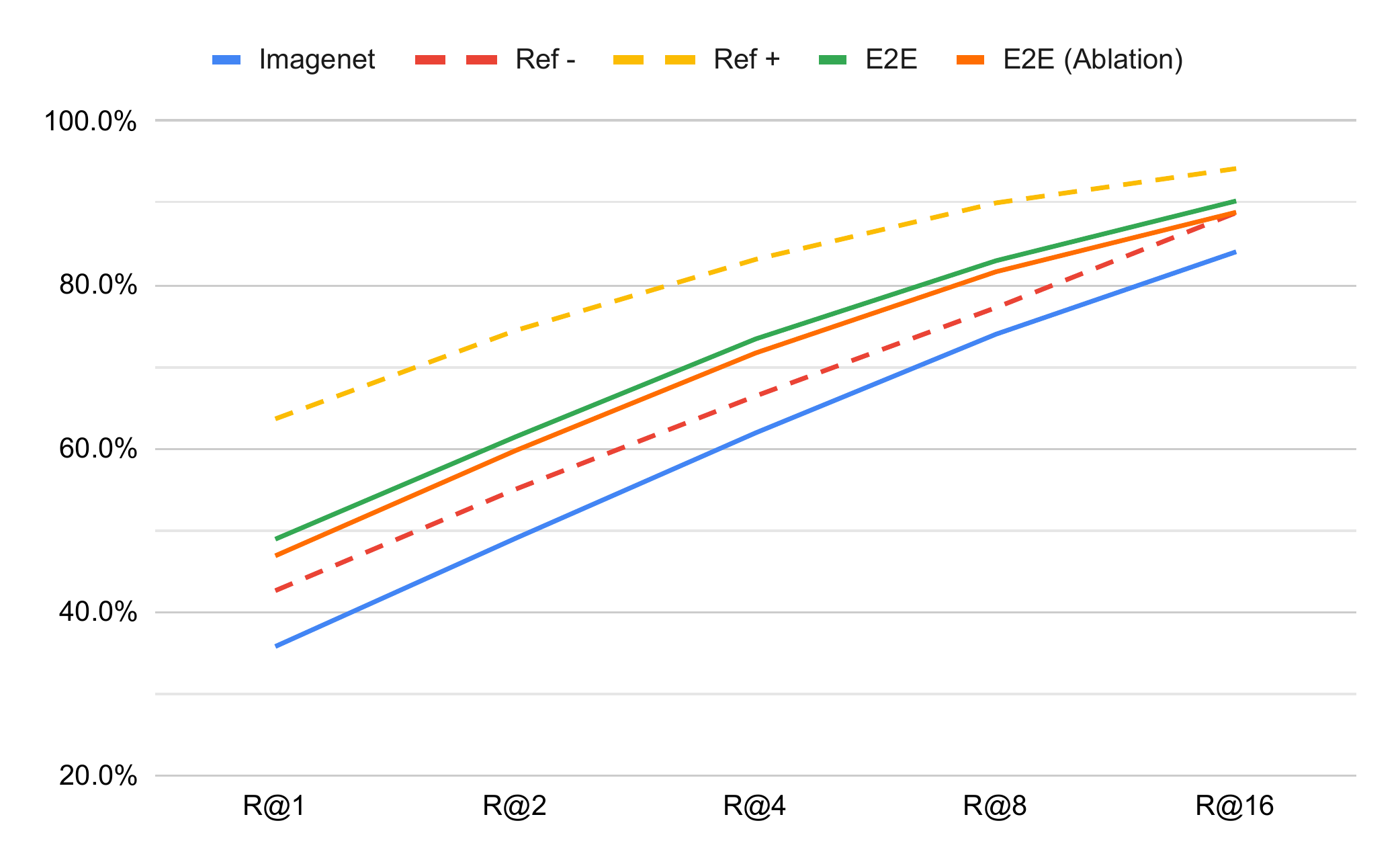}}
	\quad
	\subfloat[fig:retrieval_cars][R@k - CARS.]{\includegraphics[width=0.48\textwidth]{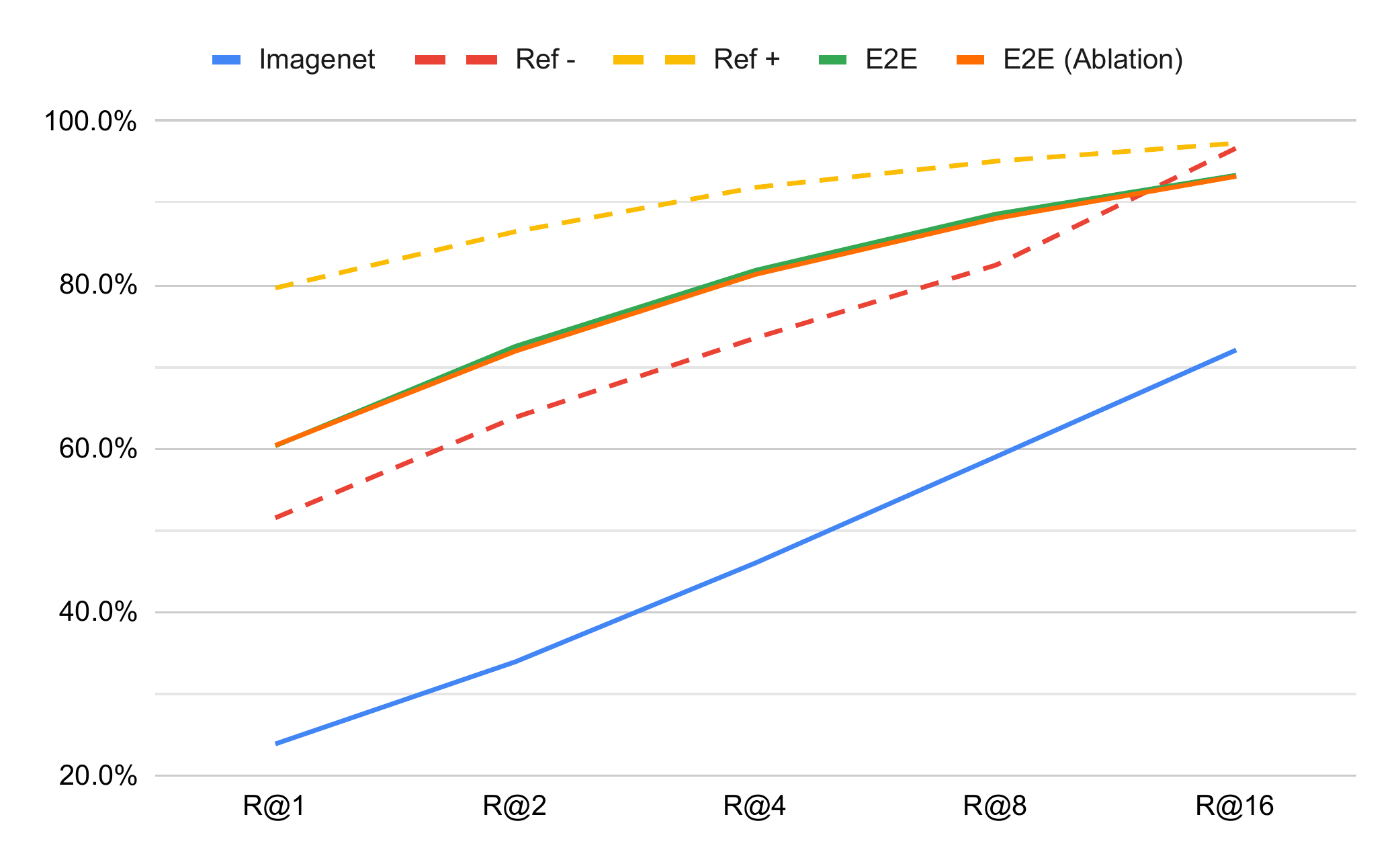}}
    \quad
	\subfloat[fig:retrieval_sop][R@k - SOP.]{\includegraphics[width=0.48\textwidth]{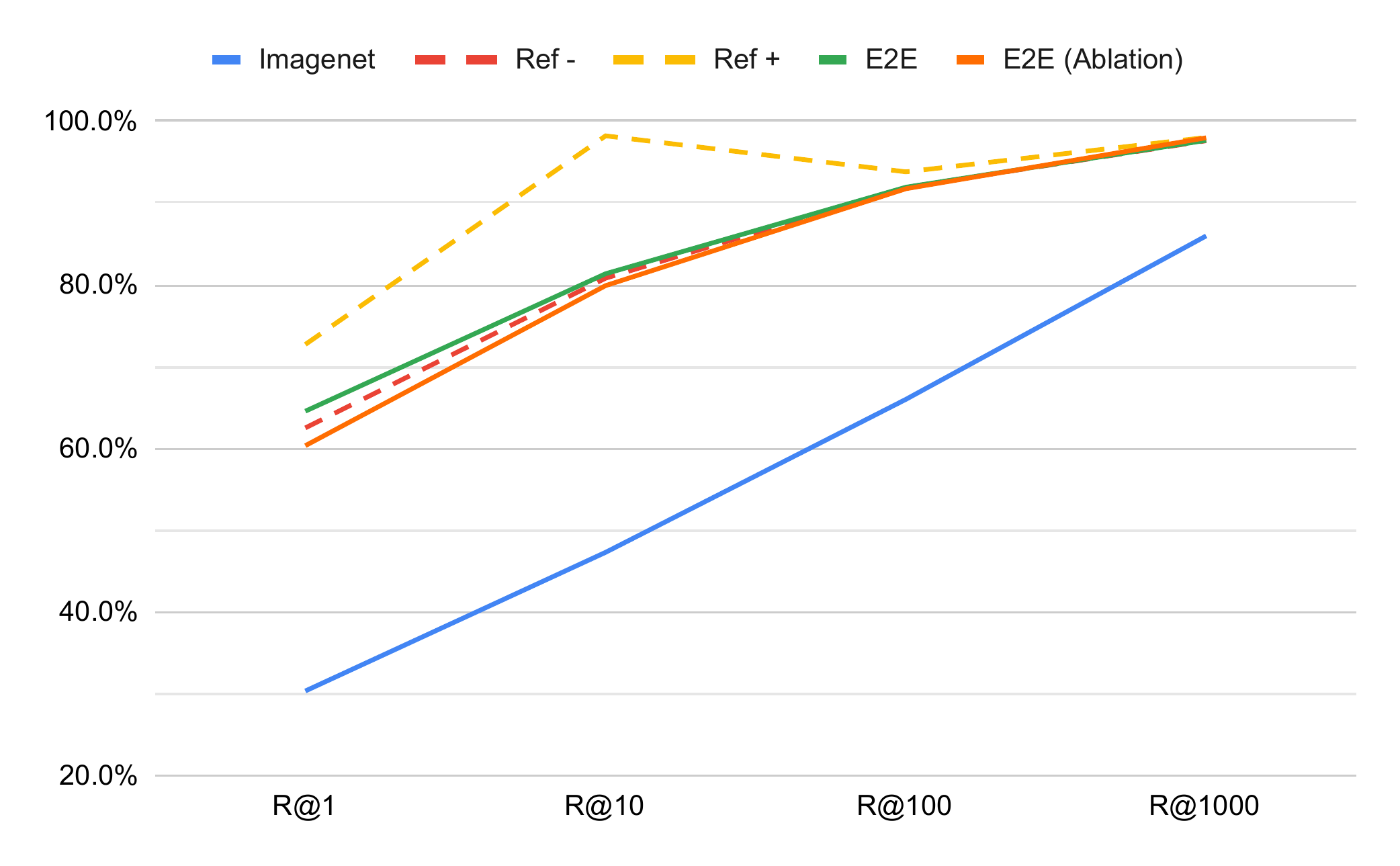}}
	\caption{Evaluation on image retrieval.}
	\label{fig:retrieval}
\end{figure}

We follow the experimental setting in past work and fine-tune pretrained models on ImageNet in each of the three datasets. The pretrained models thus correspond to the 5-layered case reported in Fig. \ref{fig:imagenet}. Fine-tuning is performed in each dataset with the same strategy to sample examples to form minibatches as reported in ImageNet experiments, while the learning rate schedule matches that of \cite{vaswani2017attention}. Results in terms of Recall@k \cite{oh2016deep} for increasing k are presented in Figures \ref{fig:retrieval}-a, \ref{fig:retrieval}-b, and \ref{fig:retrieval}-c for the cases of CUB, CARS, and SOP, respectively, while clustering performance is reported in Table \ref{tab:nmi}. We compare our models against results reported by \citet{wu2017sampling} corresponding to several metric learning schemes employed for retrieval. We thus indicate by \textit{REF. -} and \textit{REF. +} the worst and best performances they report for each metric/dataset. We further report the performance of the models trained only on ImageNet as well as an ablation case in which the auxiliary loss is dropped. In most cases our models performance lies in between \textit{REF. -} and \textit{REF. +}, i.e. a competitive performance with respect to settings heavily engineered for each case is obtained with our models where a much simpler training/inference workflow is used.

Clustering results reported in Table \ref{tab:nmi} correspond to the normalized mutual information (NMI) measured between class labels and cluster assignments obtained by the k-means algorithm executed over the representations given by $\mathcal{E}$. For our systems, we report a further result in which case a heuristic approach is used to enable the use of $\D$ to assign clusters to each data point. Using said approach, we assign each data point to the cluster corresponding to the Euclidean centroid corresponding to the smallest distance given by $\D$.

\begin{table}[h]
\centering
\caption{Clustering performance in terms of NMI. Results in parenthesis indicate the clustering performance obtained by using $\D$ to assign clusters to test examples.\label{tab:nmi}}
\begin{tabular}{cccc}
\cline{2-4}
                        & \textbf{CUB}    & \textbf{CARS}   & \textbf{SOP}    \\ \hline
\textit{ImageNet}       & 52.5\% (53.9\%) & 52.5\% (35.5\%) & 81.6\% (79.9\%) \\ \hline
\textit{Ref. -} \cite{wu2017sampling}         & 55.4\%          & 53.4\%          & 88.10\%         \\
\textit{Ref. +} \cite{wu2017sampling}         & 69.0\%          & 69.1\%          & 90.70\%         \\ \hline
\textit{E2E}            & 60.5\% (64.1\%) & 59.9\% (63.5\%) & 89.2\% (92.9\%) \\
\textit{E2E (Ablation)} & 58.5\% (62.4\%) & 60.4\% (63.3\%) & 88.1\% (92.7\%) \\ \hline
\end{tabular}
\end{table}

\end{document}